%% file: main.tex
\documentclass[final]{cvpr}

\usepackage{times}
\usepackage{epsfig}
\usepackage{graphicx}
\usepackage{amsmath}
\usepackage{amssymb}
\usepackage{amsthm}
\usepackage{booktabs}
\usepackage{bm}
\usepackage{caption}
\usepackage{xcolor}

\usepackage[pagebackref=true,breaklinks=true,colorlinks,bookmarks=false]{hyperref}

\def\cvprPaperID{} 
\def\confYear{CVPR 2021}
\newcommand{\myparagraph}[1]{\vspace{0.2em}\noindent\textbf{#1}}

\begin{document}

\title{We are More than Our Joints: Predicting how 3D Bodies Move}
\author{
  Yan Zhang$^{1}$,\; Michael J. Black$^2$,\; Siyu Tang$^{1}$ \\
  $^1$ ETH Z\"{u}rich, Switzerland \\
  $^2$Max Planck Institute for Intelligent Systems, T\"{u}bingen, Germany 
}
\twocolumn[{%
\renewcommand\twocolumn[1][]{#1}%
\maketitle
\thispagestyle{empty}
\begin{center}
    \newcommand{\teaserwidth}{\textwidth}
\vspace{-0.3in}
    \centerline{
    \includegraphics[width=0.98\textwidth]{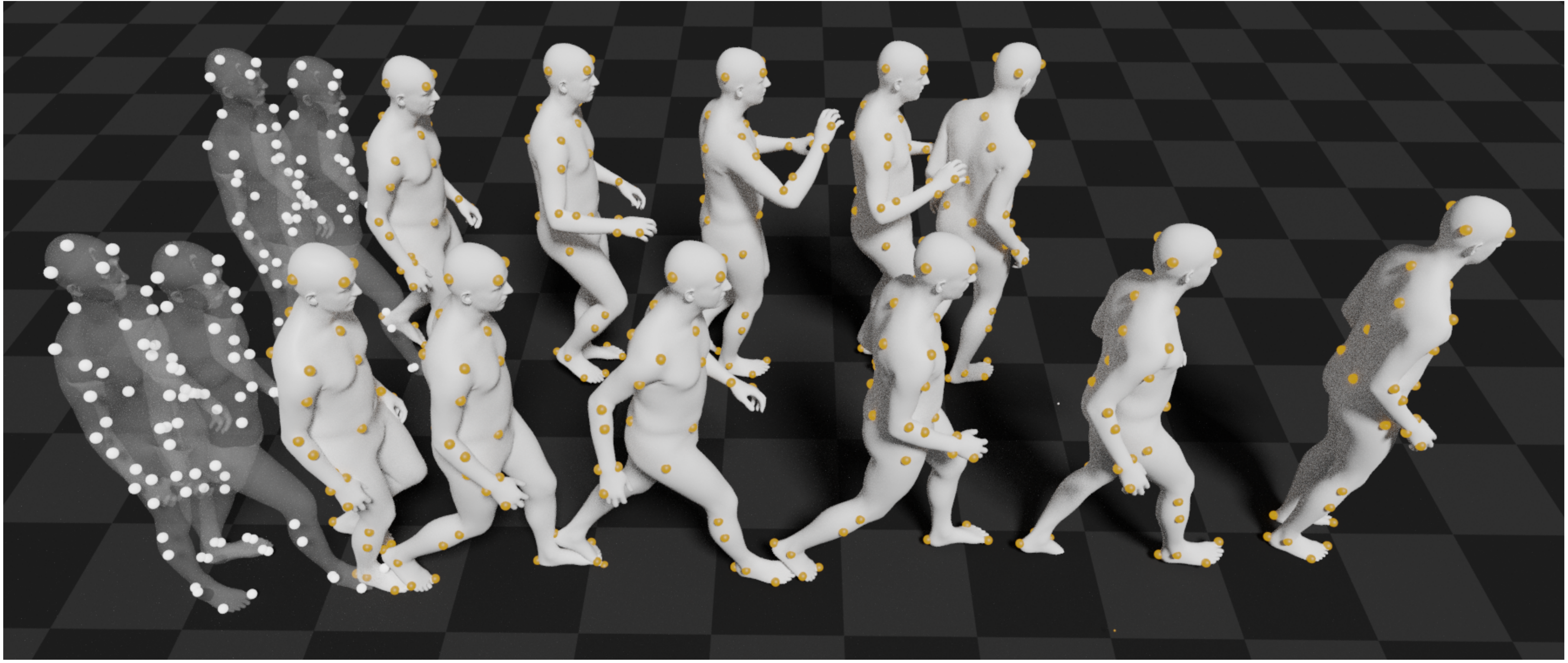}
     }
     \vspace{-0.1in}
  \captionof{figure}{We represent the body in motion with a set of 3D markers on the body surface. Given a time sequence of markers from the past (white), MOJO predicts diverse marker sequences in the future (orange) with 3D bodies they represent (gray). }  
\label{fig:teaser}
\vspace{-0.1in}
\end{center}%
}]

\maketitle

\begin{abstract}
   A key step towards understanding human behavior is the prediction of 3D human motion.
   Successful solutions have many applications in human tracking, HCI, and graphics.
   Most previous work focuses on predicting a time series of future 3D joint locations given a sequence 3D joints from the past.
   This Euclidean formulation generally works better than predicting pose in terms of joint rotations.
   Body joint locations, however, do not fully constrain 3D human pose, leaving degrees of freedom (like rotation about a limb) undefined. 
   Note that 3D joints can be viewed as a sparse {\em point cloud}. 
   Thus the problem of human motion prediction can be seen as a problem of point cloud prediction.
   With this observation, we instead predict a sparse set of locations on the body {\em surface} that correspond to motion capture markers.
   Given such markers, we fit a parametric body model to recover the 3D body of the person. 
   These sparse surface markers also carry detailed information about human movement that is not present in the joints, increasing the naturalness of the predicted motions. 
   Using the AMASS dataset, we train MOJO (More than Our JOints), which is a novel variational autoencoder with a latent DCT space that generates motions from latent frequencies. 
   MOJO preserves the full temporal resolution of the input motion, and sampling from the latent frequencies explicitly introduces high-frequency components into the generated motion.
   We note that motion prediction methods accumulate errors over time, resulting in joints or markers that diverge from true human bodies.
   To address this, we fit the SMPL-X body model to the predictions at each time step, projecting the solution back onto the space of valid bodies, before propagating the new markers in time.
   Quantitative and qualitative experiments show that our approach produces state-of-the-art results and realistic 3D body animations.
   The code is available for research purposes at {\small\url{https://yz-cnsdqz.github.io/MOJO/MOJO.html}}.
\end{abstract}


\vspace{-5mm}
\input{sec1-introduction}

\input{sec2-relatedwork}

\input{sec3-method}

\input{sec4-experiment}

\input{sec5-conclusion}

{\small
\bibliographystyle{ieee_fullname}
\bibliography{egbib}
}

\input{sec6-appendix}

\end{document}

%% file: sec1-introduction.tex
\section{Introduction}

Human motion prediction has been extensively studied as a way to understand and model human behavior.
Provided the recent past motion of a person, the goal is to predict either a deterministic or diverse set of plausible motions in the near future.
While useful for animation, AR, and VR, predicting human movement is much more valuable because it means we have a {\em model} of how people move. 
Such a model is useful for applications in sports~\cite{zhang2019predicting}, pedestrian tracking~\cite{rudenko2020human}, smart user interfaces~\cite{wu2020futurepong}, robotics~\cite{liu2017human} and more.
While this is a variant of the well-studied time-series prediction problem, most existing methods are still not able to produce realistic 3D body motion.

To address the gap in realism, we make several novel contributions but start with a few  observations.
First, most existing methods for 3D motion prediction treat the body as a skeleton and predict a small set of 3D joints.
While some methods represent the skeleton in terms of joint angles, the most accurate methods simply predict the 3D joint locations in Euclidean space.
Second, given a sparse set of joint locations, animating a full 3D body is ambiguous because important degrees of freedom are not modeled, e.g.~rotation about limb axes.
Third, most papers show qualitative results by rendering skeletons and these often look fine to the human eye.
We show, however, that, as time progresses, the skeletons can become less and less human in proportion so that, at the end of the sequence, the skeleton rarely corresponds to a valid human body.
Fourth, the joints of the body cannot capture the nuanced details of how the surface of the body moves, limiting realism of any resulting animation.

We address these issues with
a solution, called {\bf MOJO (More than Our JOints)} the predicts realistic 3D body motion.
MOJO incorporates a novel representation of the body in motion, a novel motion generative network, and a novel scheme for 3D body mesh recovery.

\textit{First}, the set of 3D joints predicted by existing methods can be viewed as a sparse {\em point cloud}.
In this light, existing human motion prediction methods preform {\em point cloud prediction}.
Thus, we are free to choose a different point cloud that better satisfies the ultimate goal of animating bodies.
Specifically, we model the body with a sparse set of surface markers corresponding to those used in motion capture (mocap) systems.
We simply swap one type of sparse point cloud for another, but, as we will show,  predicting surface markers has key advantages.
For example, there exist methods to fit a SMPL body model \cite{SMPL:2015} to such makers, producing realistic animations \cite{loper2014mosh,AMASS:ICCV:2019}.
Consequently this shift to predicting makers enables us to (1) leverage a powerful statistical body shape model to improve results, (2) immediately gives us realistic animations, (3) provides an output representation that can be used in many applications.

\textit{Second}, to model fine-grained and high-frequency interactions between markers, we design a conditional variational autoencoder (CVAE) with a latent cosine space. It not only performs stochastic motion prediction, but also improves motion realism by incorporating high-frequency motion details.
Compared to most existing methods that encode motion with a single vector (e.g.~the last hidden state of an RNN), our model preserves full temporal resolution of the sequence, and decomposes motion into independent frequency bands in the latent space via a discrete cosine transform (DCT).
Based on the energy compaction property of the DCT~\cite{ahmed1974discrete,rao2014discrete}\footnote{Most information will be concentrated at low frequency bands.}, we train our CVAE with a robust Kullback–Leibler divergence (KLD) term~\cite{zhang2020perpetual}, creating an implicit latent prior that carries most of the information at low frequency bands. 
To sample from this implicit latent prior, we employ diversifying latent flows (DLows)~\cite{yuan2020dlow} in low-frequency bands to produce informative features, and from the standard normal distribution in high-frequency bands to produce white noise. Pieces of information  from various frequencies are then fused to compose the output motion.

\textit{Third}, in the inference phase, we propose a recursive projection scheme supported by our marker-based representation, in order to retain natural body shape and pose throughout the sequence.
We regard the valid body space as a low-dimensional manifold in the Euclidean space of markers. When the CVAE decoder makes a prediction step, the predicted markers tend to leave this manifold because of error accumulation. Therefore, after each step we project the predicted markers back to the valid body manifold, by fitting an expressive SMPL-X~\cite{SMPL-X:2019} body mesh to the markers. 
On the fitted body, we know the true marker locations and pass these to the next stage of the RNN, effectively denoising the markers at each time instant.
Besides keeping the solution valid, the recursive projection scheme directly yields body model parameters and hence realistic body meshes.

We exploit the {\bf AMASS}~\cite{AMASS:ICCV:2019} dataset for evaluation, as well as {\bf Human3.6M}~\cite{h36m_pami} and {\bf HumanEva-I}~\cite{sigal2010humaneva} to compare our methods with the state-of-the-art in stochastic motion prediction.
We show that our models with the latent DCT space outperform the state-of-the-art, and that the recursive projection scheme effectively eliminates unrealistic body deformation. 
We also evaluate realism of the generated motion with a foot skating measure and a perceptual study.
Finally, we compare different body representations, in particular our solution with a traditional pipeline, which first predicts 3D joints and then fits a body to the joints. 
We find that they are comparable w.r.t.~prediction diversity and accuracy, but the traditional pipeline can produce invalid body shapes. 

\myparagraph{Contributions.}
In summary, our contributions are: 
(1) We propose a marker-based representation for bodies in motion, which provides more constraints than the body skeleton and hence benefits 3D body recovery. 
(2) We design a new CVAE with a latent DCT space to improve motion modelling. 
(3) We propose a recursive projection scheme to preserve valid bodies at test time. 

%% file: sec2-relatedwork.tex
\section{Related Work}

\textbf{Deterministic human motion prediction.}
Given an input human motion sequence, the goal is to forecast a deterministic future motion, which is expected to be close to the ground truth. This task has been extensively studied~\cite{aksan2020attention,aksan2019structured,cai2020learning,cui2020learning,ghosh2017learning,gopalakrishnan2019neural,gui2018adversarial,gui2018few,li2018convolutional,li2020dynamic,mao2019learning,martinez2017human,pavllo2019modeling,pavllo2018quaternet,tang2018long,wei2020his,zhou2018autoconditioned}.
Martinez et al.~\cite{martinez2017human} propose an RNN with residual connections linking the input and the output, and design a sampling-based loss to compensate for prediction errors during training.
Cai et al.~\cite{cai2020learning} and Mao et al.~\cite{mao2019learning} use the discrete cosine transform (DCT) to convert the motion into the frequency domain. Then Mao et al.~\cite{mao2019learning} employ graph convolutions to process the frequency components, whereas Cai et al.~\cite{cai2020learning} use a transformer-based architecture.

\textbf{Stochastic 3D human motion prediction.}
In contrast to deterministic motion prediction, stochastic motion prediction produces diverse plausible future motions, given a single motion from the past~\cite{barsoum2018hp,bhattacharyya2018accurate,dilokthanakul2016deep,gurumurthy2017deligan,ling2020character,walker2017pose,yan2018mt,yuan2020dlow,zhang2020perpetual}.
Yan et al.~\cite{yan2018mt} propose a motion transformation VAE to jointly learn the motion mode feature and transition between motion modes.
Barsoum et al.~\cite{barsoum2018hp} propose a probabilistic sequence-to-sequence model, which is trained with a Wasserstein generative adversarial network.
Bhattacharyya et al.~\cite{bhattacharyya2018accurate} design a `best-of-many' sampling objective to boost the performance of conditional VAEs.
Gurumurthy et al.~\cite{gurumurthy2017deligan} propose a GAN-based network and parameterize the latent generative space as a mixture model.
Yuan et al.~\cite{yuan2020dlow} propose diversifying latent flows (DLow) to exploit the latent space of an RNN-based VAE, which generates highly diverse but accurate future motions.

\textbf{Frequency-based motion analysis.}
Earlier studies like \cite{ormoneit2000learning,889031} adopt a Fourier transform for motion synthesis and tracking.
Akhter et al.~\cite{akhter2012bilinear} propose a linear basis model for spatiotemporal motion regularization, and discover that the optimal PCA basis of a large set of facial motion converges to the DCT basis.
Huang et al.~\cite{huang2017towards} employ low-frequency DCT bands to regularize motion of body meshes recovered from 2D body joints and silhouettes. 
The studies of~\cite{cai2020learning,mao2019learning,wei2020his} use deep neural networks to process DCT frequency components for motion prediction.
Yumer et al.~\cite{yumer2016spectral} and Holden et al.~\cite{holden2017fast} handle human motions in the Fourier domain to conduct motion style transfer.

\textbf{Representing human bodies in motion.}
3D joint locations are widely used, e.g.~\cite{li2020dynamic,martinez2017human,yuan2020dlow}. 
To improve prediction accuracy, Mao et al.~\cite{mao2019learning}, Cui et al.~\cite{cui2020learning}, Li et al.~\cite{li2020dynamic} and others use a graph to capture interaction between joints.
Askan et al.~\cite{aksan2019structured} propose a structured network layer to represent the body joints according to a kinematic tree.
Despite their effectiveness, the skeletal bone lengths can vary during motion.
To alleviate this issue, Hernandez et al.~\cite{hernandez2019human} use a training loss to penalize bone length variations. 
Gui et al.~\cite{gui2018adversarial} design a geodesic loss and two discriminators to keep the predicted motion human-like over time.
To remove the influence of body shape, Pavllo et al.~\cite{pavllo2019modeling,pavllo2018quaternet} use quaternion-based joint rotations to represent the body pose. 
When predicting the global motion, a walking path is first produced and the pose sequence is then generated. 
Zhang et al.~\cite{zhang2020perpetual} represent the body in motion by the 3D global translation and the joint rotations following the SMPL kinematic tree~\cite{SMPL:2015}. When animating a body mesh, a constant body shape is added during testing.
Although a constant body shape is preserved, foot skating frequently occurs due to the inconsistent relation between the body pose, the body shape and the global translation.

\textbf{MOJO in context.}
Our solution not only improves stochastic motion prediction over the state-of-the-art, but also directly produces diverse future motions of realistic 3D bodies. 
Specifically, our latent DCT space represents motion with different frequencies, rather than a single vector in the latent space. We find that generating motions from different frequency bands significantly improves diversity while retaining accuracy.
Additionally, compared to previous skeleton-based body representations, we propose to use markers on the body surface to provide more constraints on the body shape and DoFs. 
This marker-based representation enables us to design an efficient recursive projection scheme by fitting SMPL-X~\cite{SMPL-X:2019} at each prediction step. 
Recently, in the context of autonomous driving, Weng et al.~\cite{Weng2020_SPF2_eccvw,Weng2020_SPF2} forecast future LiDAR point clouds and then detect-and-track 3D objects in the predicted clouds. 
While this has similarities to MOJO, they do not address articulated human movements.

%% file: sec3-method.tex
\section{Method}
\label{sec:method}

\subsection{Preliminaries}

\myparagraph{SMPL-X body mesh model~\cite{SMPL-X:2019}.}
Given a compact set of body parameters, SMPL-X produces a realistic body mesh including face and hand details. 
In our work, the body parameter set ${\bm \Theta}$ includes the global translation ${\bm t} \in \mathbb{R}^3$, the global orientation ${\bm R}\in\mathbb{R}^6$ w.r.t. the continuous representation~\cite{zhou2019continuity}, the body shape ${\bm \beta}\in\mathbb{R}^{10}$, the body pose ${\bm \theta}\in\mathbb{R}^{32}$ in the VPoser latent space~\cite{SMPL-X:2019}, and the hand pose ${\bm \theta^h}\in\mathbb{R}^{24}$ in the MANO~\cite{MANO:SIGGRAPHASIA:2017} PCA space. 
We denote a SMPL-X mesh as $\mathcal{M}({\bm \Theta})$, which has a fixed topology with 10,475 vertices.
MOJO is implemented using SMPL-X in our work, but any other parametric 3D body model could be used, e.g.~\cite{STAR:ECCV:2020,xu2020ghum}.

\myparagraph{Diversifying latent flows (DLow)~\cite{yuan2020dlow}.}
The entire DLow method comprises a CVAE to predict future motions, and a network \textbf{Q} that takes the condition sequence as input and transforms a sample $\varepsilon \sim \mathcal{N}(0, {\bm I})$ to $K$ diverse places in the latent space.
To train the CVAE, the loss consists of a frame-wise reconstruction term, a term to penalize the difference between the last input frame and the first predicted frame, and a KLD term. 
Training the network \textbf{Q} requires a pre-trained CVAE decoder, their training loss encourages diverse generated motion samples in which, at least one sample is close to the ground truth motion.

\subsection{Human Motion Representation}
\label{sec:motion_representation}
Most existing methods use 3D joint locations or rotations to represent the body in motion. This results in ambiguities in recovering the full shape and pose \cite{Bogo:ECCV:2016, SMPL-X:2019}.
To obtain more constraints on the body, while preserving efficiency, we represent bodies in motion with markers on the body surface. 
Inspired by modern mocap systems, we follow the marker placements of either the {CMU} mocap dataset~\cite{mocap_cmu} or the SSM2 dataset~\cite{AMASS:ICCV:2019}, and select V vertices on the SMPL-X body mesh, which are illustrated in the Appendix.

The 3D markers are simply points in Euclidean space.
Compared to representing the body with the global translation and the local pose, like in~\cite{pavllo2018quaternet,zhang2020perpetual}, such a location-based representation naturally couples the global body translation and the local pose variation, and hence is less prone to motion artifacts like foot skating, which are caused by the mismatch between the global movement, the pose variation, and the body shape.

Therefore, in each frame the body is represented by a V-dimensional feature vector, i.e.~the concatenation of the 3D locations of the markers, and the motion is represented by a time sequence of such vectors.
We denote the input sequence to the model as ${\bm X}:=\{{\bm x}_t\}_{t=0}^M$, and a predicted future sequence from the model as ${\bm Y}:=\{{\bm y}_t\}_{t=0}^N$, where ${\bm y}_0 = {\bm x}_{M+1}$. 
With fitting algorithms like MoSh and MoSh++~\cite{loper2014mosh,AMASS:ICCV:2019}, it is much easier to recover 3D bodies from markers than from joints.


\subsection{Motion Generator with Latent Frequencies}
\label{sec:generative_model}

\begin{figure}
    \centering
    \includegraphics[width=\linewidth]{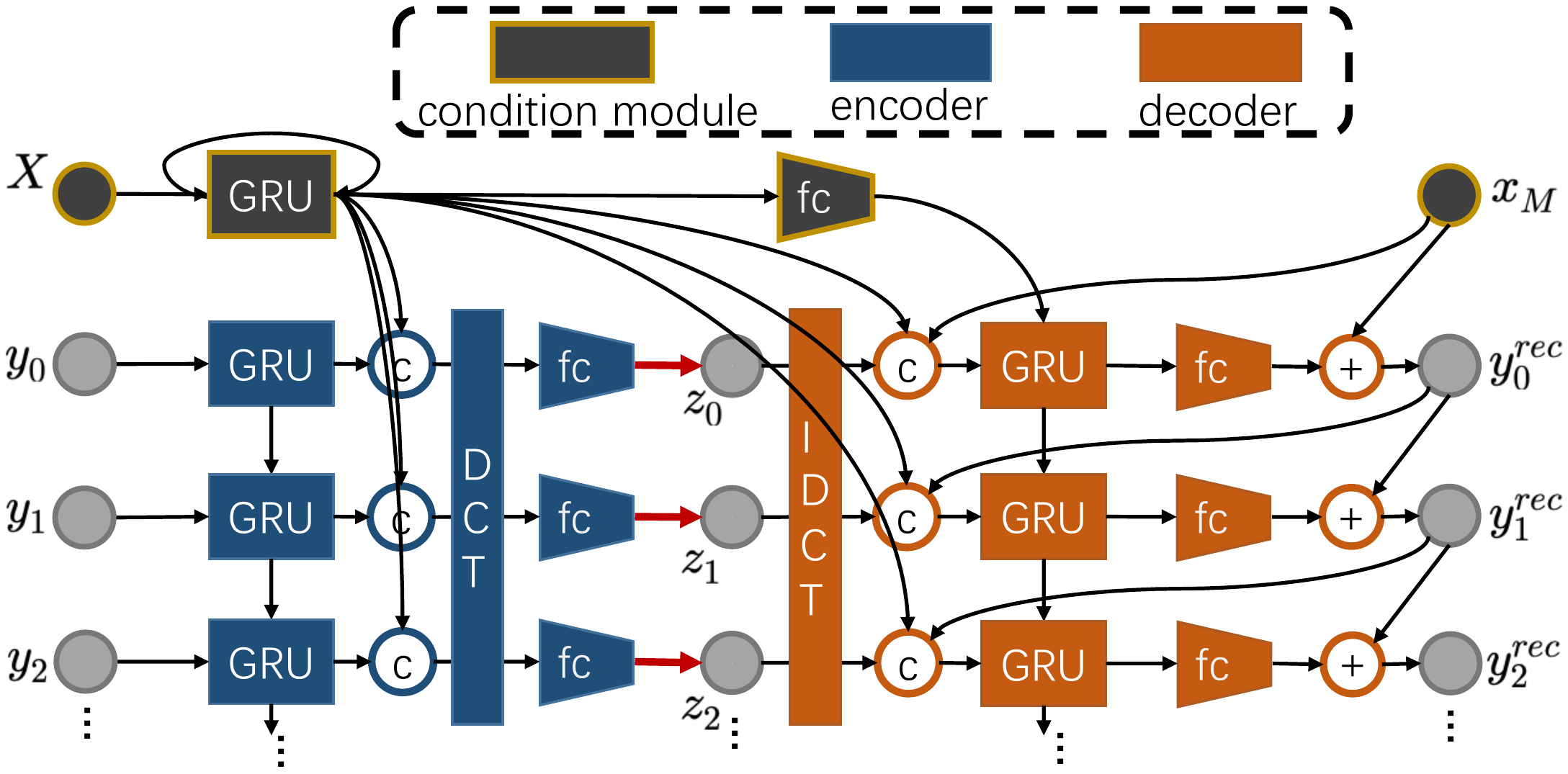}
     \vspace{-0.25in}
   \caption{Illustration of our CVAE architecture. The red arrows denote sampling from the inference posterior. The circles with `c' and `+' denote feature concatenation and addition, respectively. The blocks with `fc' denote a stack of fully-connected layers.}
    \label{fig:architectures}
\end{figure}

For a real human body, the motion granularity usually corresponds to the motion frequency because of the underlying locomotor system. For example, the frequency of waving hands is usually much higher than the frequency of a jogging gait. 
Therefore, we design a network with multiple frequency bands in the latent space, so as to better represent interactions between markers on the body surface and to model motions at different granularity levels.

\myparagraph{Architectures.}
Our model is visualized in Fig.~\ref{fig:architectures}, which is designed according to the CVAE in the DLow method~\cite{yuan2020dlow}.
The encoder with a GRU~\cite{cho-etal-2014-learning} preserves full temporal resolution of the input. Then, the motion information is decomposed onto multiple frequency bands via DCT. 
At individual frequency bands, we use the re-parameterization trick~\cite{kingma2013auto} to introduce randomness, and then use inverse DCT to convert the motion back to the temporal domain. To improve temporal smoothness and eliminate the first-frame jump reported in \cite{martinez2017human}, we use residual connections at the output. 
We note that the CVAE in the DLow method, which does not have residual connections but has a loss to penalize the jump artifact, does not produce smooth and realistic marker motions.
A visualization of this latent DCT space is shown in the Appendix.

\myparagraph{Training with robust Kullback-Leibler divergence.}
Our training loss comprises three terms for frame-wise reconstruction, frame-wise velocity reconstruction, and latent distribution regularization, respectively;
\begin{equation}
    \begin{split}
    \mathcal{L} &= \mathbb{E}_{{\bm Y}}[|{\bm Y}-{\bm Y}^{rec}|] + \alpha \mathbb{E}_{{\bm Y}}[|\Delta~{\bm Y}-\Delta~{\bm Y}^{rec}|] \\ &+ \Psi\left(KLD( q({\bm Z} | {{\bm X},{\bm Y}}) || \mathcal{N}(0, {\bm I}))) \right) ,
    \end{split}
    \label{eq:trainloss}
\end{equation}
where the operation $\Delta$  computes the time difference, $q(\cdot | \cdot )$ denotes the inference posterior (the encoder), ${\bm Z}$ denotes the latent frequency components, and $\alpha$ is a loss weight.
We find the velocity reconstruction term can further improve temporal smoothness. 

Noticeably, our distribution regularization term is given by the robust KLD~\cite{zhang2020perpetual} with $\Psi(s) = \sqrt{1+s^2}-1$~\cite{charbonnier1994two}, which defines an implicit latent prior different from the standard normal distribution.
During optimization, the gradients to update the entire KLD term become smaller when the divergence from $\mathcal{N}(0, {\bm I})$ becomes smaller.
Thus, it expects the inference posterior to carry information, and prevents posterior collapse.
More importantly, this term is highly suitable for our latent DCT space. 
According to the energy compaction property of DCT \cite{ahmed1974discrete,rao2014discrete}, we expect that the latent prior deviates from $\mathcal{N}(0, {\bm I})$ at low-frequency bands to carry information, but equals $\mathcal{N}(0, {\bm I})$ at high-frequency bands to produce white noise.
We let this robust KLD term determine which frequency bands to carry information automatically.

\myparagraph{Sampling from the latent DCT space.}
Since our latent prior is implicit, sampling from the latent space is not as straightforward as sampling from the standard normal distribution, like for most VAEs. 
Due to the DCT nature, we are aware that motion information is concentrated at low-frequency bands, and hence we directly explore these informative low-frequency bands using the network \textbf{Q} in DLow~\cite{yuan2020dlow}.

Specifically, we use $\{\textbf{Q}_w\}_{w=1}^L$ to sample from the lowest $L$ frequency bands, and sample from $\mathcal{N}(0, {\bm I})$ from $L+1$ to the highest frequency bands. Since the cosine basis is orthogonal and individual frequency bands carry independent information, these $L$ DLow models do not share parameters, but are jointly trained with the same losses as in \cite{yuan2020dlow}, as well as the decoder of our CVAE with the latent DCT space.
Our sampling approach is illustrated in Fig.~\ref{fig:fdlow}. 
The influence of the threshold $L$ is investigated in Sec.~\ref{sec:experiments} and in the Appendix.

\begin{figure}
    \centering
    \includegraphics[width=\linewidth]{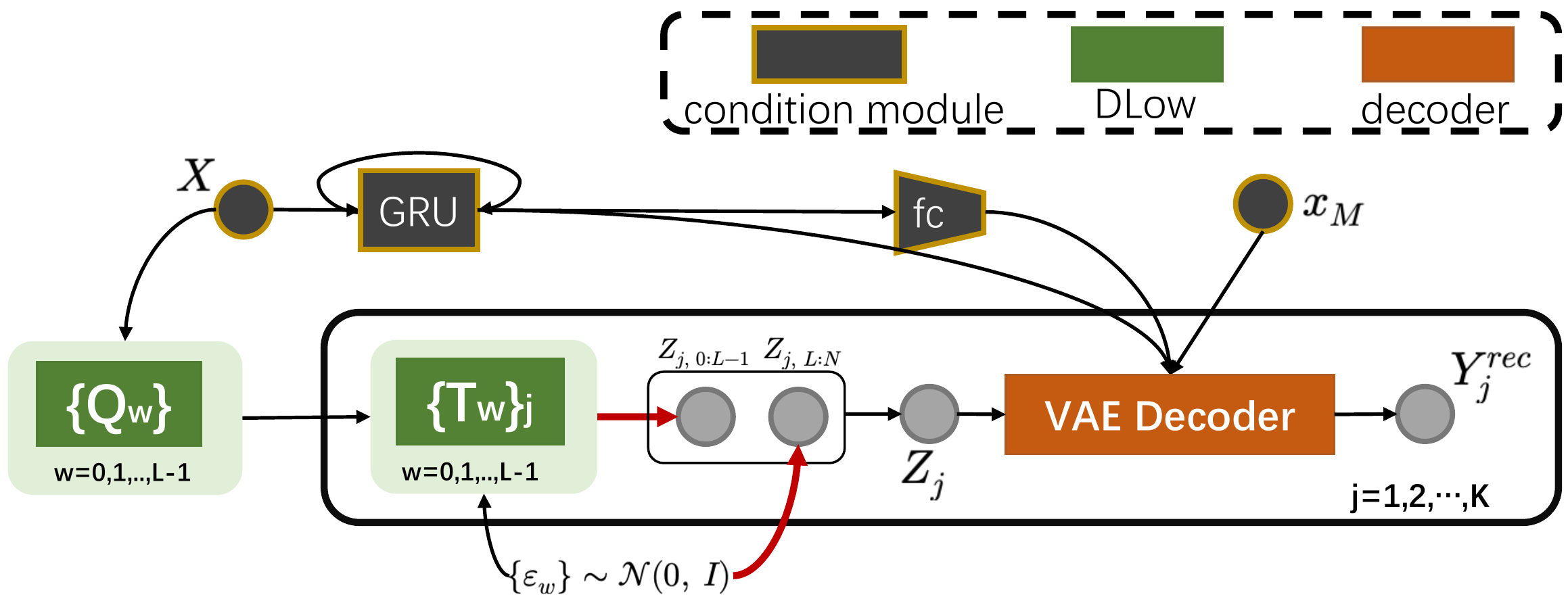}
    \vspace{-0.25in}
    \caption{Illustration of our sampling scheme to generate $K$ different sequences. ${\bm Q}_w$ denotes the network to produce the latent transformation ${\bm T}_w$ at the frequency band $w$. The red arrows denote the sampling operation.}
    \label{fig:fdlow}
\end{figure}

\subsection{Recursive Projection to the Valid Body Space}

\begin{figure}
    \centering
    \includegraphics[width=\linewidth]{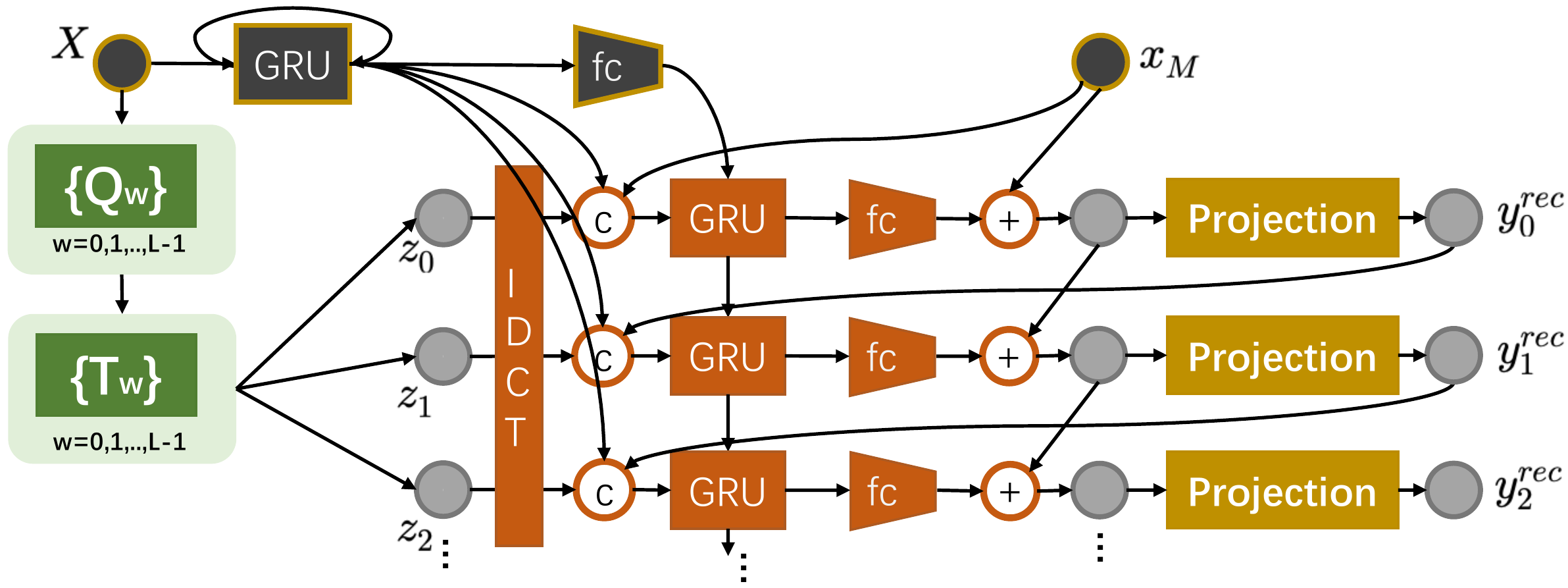}
    \vspace{-0.25in}
    \caption{Illustration of our prediction scheme with projection. The notation has the same meaning as before.}
    \label{fig:fitting}
\end{figure}

Our generative model produces diverse motions in terms of marker location variations. 
Due to RNN error accumulation, predicted markers can gradually deviate from a valid 3D body, resulting in, e.g., flattened heads and twisted torsos.
Existing methods with new losses or discriminators ~\cite{gui2018adversarial,hernandez2019human,VIBE:CVPR:2020} can alleviate this problem, but may unpredictably fail due to the train-test domain gap.

Instead, we exploit the fact that valid bodies lie on a low-dimensional manifold in the Euclidean space of markers.
Whenever the RNN performs a prediction step, the solution tends to leave this manifold.
Therefore, at each prediction step, we project the predicted markers back to that manifold, by fitting a SMPL-X body mesh to the predicted markers. 
Since markers provide rich body constraints, and we start close to the solution, the fitting process is efficiently. 
Our recursive projection scheme is illustrated in Fig.~\ref{fig:fitting}.
Note that we only apply recursive projection at the inference stage.

Following the work of MoSh~\cite{loper2014mosh} and MoSh++~\cite{AMASS:ICCV:2019}, the fitting is optimization-based, and consists of three stages: (1) optimizing the global configurations ${\bm t}$ and ${\bm R}$, (2) additionally optimizing the body pose ${\bm \theta}$, and (3) additionally optimizing the hand pose ${\bm \theta}^h$. 
At each time $t$, we use the previous fitted result to initialize the current optimization process, so that the optimum can be reached with a small number of iterations.
The loss of our optimization-based fitting at time $t$ is given by
\begin{equation}
    \mathcal{L}_f({\bm \Theta}_t):= |V_{\mathcal{M}({\bm \Theta}_t)}-{\bm y}_t^{pred}|^2 + \lambda_1 |{\bm \theta}_t|^2 + \lambda_2 |{\bm \theta}^h_t|^2,
    \label{eq:fitloss}
\end{equation}
in which $\lambda$s are the loss weights, $V$ denotes the corresponding markers on the SMPL-X body mesh, and ${\bm y}_t^{pred}$ denotes the markers predicted  by the CVAE decoder. 
The  recursive  projection  uses the body shape from the input sequence, and runs at 2.27sec/frame on average in our trials, which is comparable to the pose stage of MoSh++.
From our recursive projection scheme, we not only obtain regularized markers, but also realistic 3D bodies as well as their characteristic parameters.

%% file: sec4-experiment.tex
\section{Experiment}
\label{sec:experiments}

MOJO has several components that we evaluate.
First, to test the effectiveness of the MOJO CVAE architecture, in particular the benefits of the latent DCT spcae, we evaluate stochastic motion prediction in Sec.~\ref{sec:exp:stochastic_prediction}.
Second, to test the effectiveness of MOJO with recursive projection, we evaluate the performance of MOJO w.r.t.~realism of 3D body movements in Sec.~\ref{sec:exp:motion_realism}.
Finally, to test the advantage of the marker-based representation, we systematically compare different body representations in Sec.~\ref{sec:exp:markers_vs_jts}.
We find that MOJO produces diverse realistic 3D body motions and outperforms the state-of-the-art.


\subsection{Datasets}
For training, we use the {\bf AMASS}~\cite{AMASS:ICCV:2019} dataset. 
Specifically, we train the models on {\bf CMU}~\cite{mocap_cmu} and {\bf MPI HDM05}~\cite{muller2007documentation}, and test models on {\bf ACCAD}~\cite{accad} and {\bf BMLhandball}~\cite{helm2017motion,helm2020integrating}. 
%
This gives 696.1 minutes of training motion from 110 subjects and
128.72 minutes of test motion from 30 subjects.
The test sequences include a wide range of actions. 
To unify the sequence length and world coordinates, we canonicalize {\bf AMASS} sequences as a pre-processing step. Details are in the Appendix.

To compare our method with SOTA stochastic motion prediction methods, we additionally perform skeleton-based motion prediction on the {\bf Human3.6M} dataset~\cite{h36m_pami} and the {\bf HumanEva-I} dataset~\cite{sigal2010humaneva}, following the experimental setting of Yuan et al.~\cite{yuan2020dlow}.

\subsection{Baselines}
MOJO predicts surface markers, and has several components.
Unless mentioned, we use the CMU layout with 41 markers.
`MOJO-DCT' is the model without DCT, but with the same latent space as the CVAE in DLow. 
`MOJO-proj' is the model without  recursive projection.
Note that the suffixes can be concatenated; e.g.~`MOJO-DCT-proj' is the model without the latent DCT space and without the recursive projection scheme.

\subsection{Evaluation of Stochastic Motion Prediction}
\label{sec:exp:stochastic_prediction}
\subsubsection{Metrics}

\myparagraph{Diversity.} 
We use the same diversity measure as \cite{yuan2020dlow}, which is the average pair-wise distance between all generated sequences. 

\myparagraph{Prediction accuracy.} 
As in~\cite{yuan2020dlow}, we use the average distance error (\textit{ADE}) and the final distance error (\textit{FDE}) to measure the minimum distance between the generated motion and the ground truth, w.r.t.~frame-wise difference and the final frame difference, respectively.
Additionally, we use \textit{MMADE} and \textit{MMFED} to evaluate  prediction accuracy when the input sequence slightly changes;  see  Appendix.

\myparagraph{Motion Frequency.} Similar to \cite{hernandez2019human}, we compute the frequency spectra entropy (\textit{FSE}) to measure motion frequency in the Fourier domain, which is given by the averaged spectra entropy minus the ground truth. 
A higher value indicates the generated motions contain more motion detail. 
Note that high frequency can also indicate noise, and hence this metric is jointly considered with the prediction accuracy.

\subsubsection{Results}
We generate 50 different future sequences based on each input sequence, as in~\cite{yuan2020dlow}. 
Here we focus only on evaluating performance on motion prediction, and hence do not incorporate the body re-projection scheme.
%
The results are shown in Tab.~\ref{tab:marker_compare}, in which we employ DLow on the 20\% (i.e.~the first 9) lowest frequency bands in `MOJO-proj'.
We find that DCT consistently leads to better performance.
Noticeably, higher motion frequency indicates that the generated motions contain more details, and hence are more realistic.


\begin{table*}
    \scriptsize
    \centering
    \begin{tabular}{lllllllllllll}
        \toprule
        &\multicolumn{6}{c}{\textbf{ACCAD}~\cite{accad}} & \multicolumn{6}{c}{\textbf{BMLHandball}~\cite{helm2017motion,helm2020integrating}}\\
        \cmidrule(lr){2-7} \cmidrule(lr){8-13}
        Method & \textit{Diversity}$\uparrow$ & \textit{ADE}$\downarrow$ & \textit{FDE}$\downarrow$ & \textit{MMADE}$\downarrow$ & \textit{MMFDE}$\downarrow$ & \textit{FSE}$\uparrow$ & \textit{Diversity}$\uparrow$ & \textit{ADE}$\downarrow$ & \textit{FDE}$\downarrow$ & \textit{MMADE}$\downarrow$ & \textit{MMFDE}$\downarrow$ & \textit{FSE}$\uparrow$ \\
        \midrule
        MOJO-DCT-proj & 25.349 & \textbf{1.991} & 3.216 & 2.059 & 3.254 & 0.4 & 21.504 & 1.608 & 1.914 & 1.628 & 1.919 & 0.0\\
        MOJO-proj & \textbf{28.886} & 1.993 & \textbf{3.141} & \textbf{2.042} & \textbf{3.202} & \textbf{1.2} & \textbf{23.660} & \textbf{1.528} & \textbf{1.848} & \textbf{1.550} & \textbf{1.847} & \textbf{0.4}\\
        \bottomrule
    \end{tabular}
    \vspace{-0.1in}
    \caption{Comparison between generative models for predicting marker-based motions. The symbol $\downarrow$ (or $\uparrow$) denotes whether results that are lower (or higher) are better, respectively. Best results of each model are in boldface. The FSE scores are on the scale of $10^{-3}$. }
    \label{tab:marker_compare}
\end{table*}

To further investigate the benefits of our latent DCT space, we add the latent DCT space into the DLow CVAE model~\cite{yuan2020dlow} and train it with the robust KLD term. For sampling, we apply a set of DLow models $\{\textbf{Q}_w\}$ on the lowest $L$ bands, as in Sec.~\ref{sec:generative_model}. We denote this modified model as `VAE+DCT+$L$'. Absence of the suffix `+$L$' indicates sampling from $\mathcal{N}(0, {\bm I})$ in all frequency bands.

A comparison with existing methods is shown in Tab.~\ref{tab:skeleton_compare}.
Overall,  our latent DCT space effectively improves on the state-of-the-art. The diversity is improved by a large margin, while the prediction accuracies are comparable to the baseline.
The performance w.r.t.~\textit{MMADE} and \textit{MMFDE} is slightly inferior. 
A probable reason is VAE+DCT uses high-frequency components to generate motions, which makes motion prediction sensitive to slight changes of the input.
Moreover, by comparing `VAE+DCT' and `VAE+DCT+L', we can see that sampling from $\mathcal{N}(0, {\bm I})$ yields much worse results. This indicates that sampling from the standard normal distribution, which treats all frequency bands equally, cannot effectively exploit the advantage of the latent DCT space. Note that most information is in the low-frequency bands (see Appendix), and hence our proposed sampling method utilizes the latent frequency space in a more reasonable way and produces better results.


\begin{table*}[t!]
    \scriptsize
    \centering
    \begin{tabular}{lrccccrcccc}
        \toprule
        &\multicolumn{5}{c}{\textbf{Human3.6M}~\cite{h36m_pami}} & \multicolumn{5}{c}{\textbf{HumanEva-I}~\cite{sigal2010humaneva}}\\
        \cmidrule(lr){2-6} \cmidrule(lr){7-11}
        Method & \textit{Diversity}$\uparrow$ & \textit{ADE}$\downarrow$ & \textit{FDE}$\downarrow$ & \textit{MMADE}$\downarrow$ & \textit{MMFDE}$\downarrow$ & \textit{Diversity}$\uparrow$ & \textit{ADE}$\downarrow$ & \textit{FDE}$\downarrow$ & \textit{MMADE}$\downarrow$ & \textit{MMFDE}$\downarrow$ \\
        \midrule
        Pose-Knows~\cite{walker2017pose} & 6.723 & 0.461 & 0.560 & 0.522 & 0.569 & 2.308 & 0.269 & 0.296 & 0.384 & 0.375\\
        MT-VAE~\cite{yan2018mt} & 0.403 & 0.457 & 0.595 & 0.716 & 0.883 & 0.021 & 0.345 & 0.403 & 0.518 & 0.577\\
        HP-GAN~\cite{barsoum2018hp} & 7.214 & 0.858 & 0.867 & 0.847 & 0.858 & 1.139 & 0.772 & 0.749 & 0.776 & 0.769 \\
        Best-of-Many~\cite{bhattacharyya2018accurate} & 6.265 & 0.448 & 0.533 & 0.514 & 0.544 & 2.846 & 0.271 & 0.279 & 0.373 & 0.351\\
        GMVAE~\cite{dilokthanakul2016deep} & 6.769 & 0.461 & 0.555 & 0.524 & 0.566 & 2.443 & 0.305 & 0.345 & 0.408 & 0.410\\
        DeLiGAN~\cite{gurumurthy2017deligan} & 6.509 & 0.483 & 0.534 & 0.520 & 0.545 & 2.177 & 0.306 & 0.322 & 0.385 & 0.371\\
        DSF~\cite{yuan2019diverse} & 9.330 & 0.493 & 0.592 & 0.550 & 0.599 & 4.538 & 0.273 & 0.290 & 0.364 & 0.340\\
        DLow~\cite{yuan2020dlow} & 11.730 & 0.425 & 0.518 & \textbf{0.495} & \textbf{0.532} & 4.849 & 0.246 & 0.265 & \textbf{0.360} & \textbf{0.340} \\
        \midrule
        VAE+DCT & 3.462 & 0.429 & 0.545 & 0.525 & 0.581 & 0.966 & 0.249 & 0.296 & 0.412 & 0.445 \\
        VAE+DCT+5 & 12.579 & \textbf{0.412} & \textbf{0.514} & 0.497 & 0.538 & 4.181 & \textbf{0.234} & \textbf{0.244} & 0.369 & 0.347 \\
        VAE+DCT+20 & \textbf{15.920} & 0.416 & 0.522 & 0.502 & 0.546 & \textbf{6.266} & 0.239 & 0.253 & 0.371 & 0.345 \\
        \bottomrule
    \end{tabular}
    \vspace{-0.1in}
    \caption{Comparison between a baseline with our latent DCT space and the state-of-the-art. Best results are in boldface.  }
    \label{tab:skeleton_compare}
\end{table*}

\subsection{Evaluation of Motion Realism}
\label{sec:exp:motion_realism}

\subsubsection{Metrics}
The motion prediction metrics cannot indicate whether a motion is realistic or not. 
Here, we employ MOJO with recursive projection to obtain 3D body meshes, and evaluate the body deformation, foot skating, and perceptual quality.

\myparagraph{Body deformation.}
Body shape can be described by the pairwise distances between markers.
As a body moves, there is natural variation in these distances.
Large variations, however, indicate a deformed body that no longer corresponds to any real person.
We use variations in pairwise marker distances for the head, upper torso, and lower torso as a measure of how distorted the predicted body is.
See Appendix for the metric details.

\myparagraph{Foot skating ratio.}
Foot skating is measured based on the two markers on the lef and right foot calcaneus (`LHEE' and `RHEE' in \textbf{CMU}~\cite{mocap_cmu}). 
We consider foot skating to have occurred, when both foot markers are close enough to the ground (within 5cm) and simultaneously exceed a speed limit (5mm between two consecutive frames or 75mm/s).
We report the averaged ratio of frames with foot skating.

\myparagraph{Perceptual score.}
We render the generated body meshes as well as the ground truth, and perform a perceptual study on Amazon Mechanical Turk.
Subjects see a motion sequence and the statement ``The human motion is natural and realistic."  They evaluate this on a six-point Likert scale from `strongly disagree` (1) to `strongly agree' (6). 
Each individual video is rated by three subjects.
We report mean values and standard deviations for each method and each dataset.

\subsubsection{Results}
\begin{figure}
    \centering
    \includegraphics[width=\linewidth]{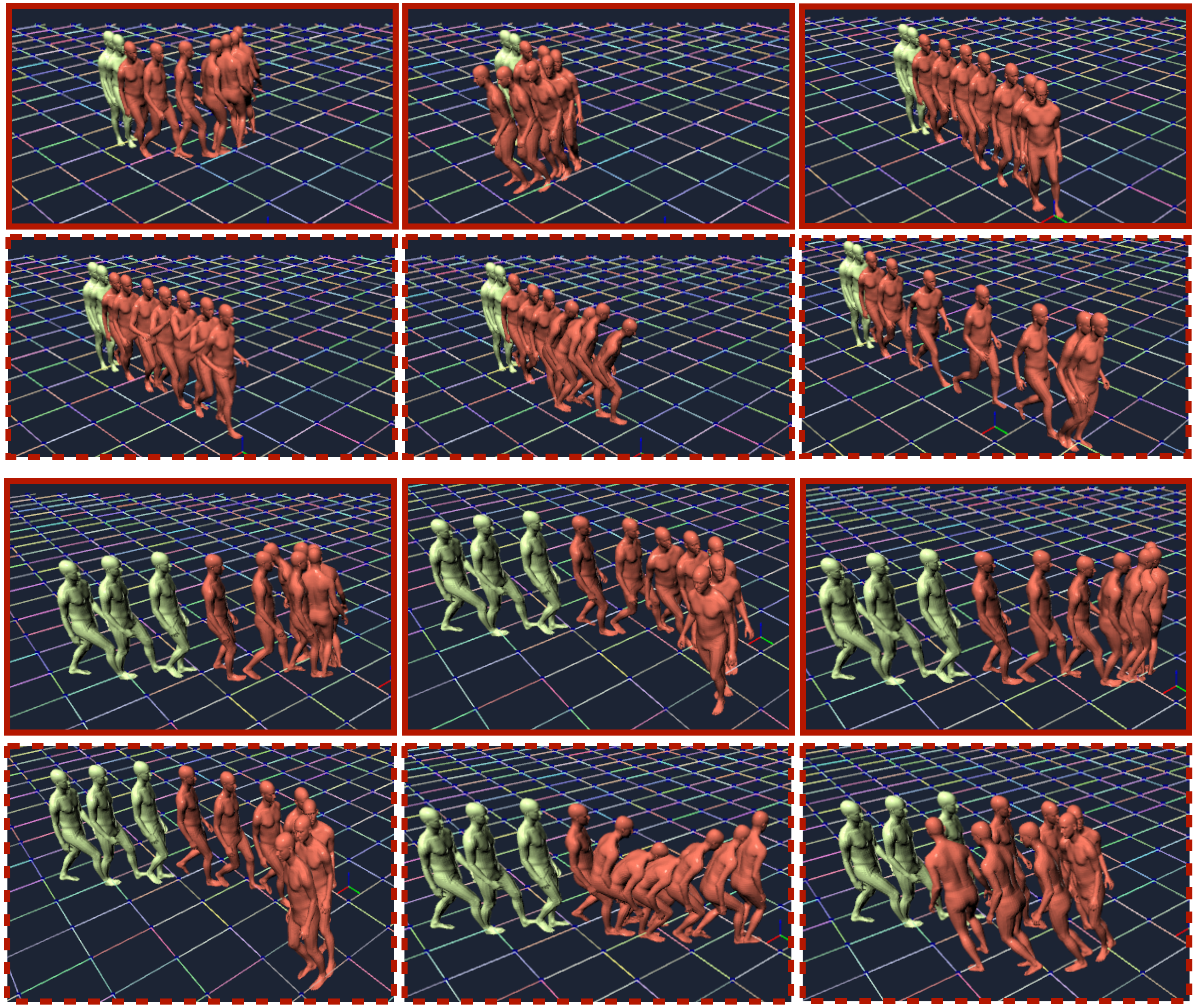}
        \vspace{-0.25in}
    \caption{Visualization of 3D body motions. Bodies in gray-green and red denote the input and the generated motion, respectively. The solid and dash image borders denote the results from MOJO-DCT and MOJO, respectively.}       
    \label{fig:bm_diverse}
\end{figure}

We randomly choose 60 different sequences from {\bf ACCAD} and {\bf BMLhandball}, respectively. Based on each sequence, we generate 15 future sequences.
Figure \ref{fig:bm_diverse} shows some generated 3D body motions. 
The motions generated by MOJO contain finer-grained body movements.

\begin{table}
    \scriptsize
    \centering
    \begin{tabular}{lcccccc}
        \toprule
          & \multicolumn{3}{c}{\textbf{ACCAD}~\cite{accad}} & \multicolumn{3}{c}{\textbf{BMLHandball}~\cite{helm2017motion,helm2020integrating}}\\
        \cmidrule(lr){2-4} \cmidrule(lr){5-7}
        Method & Head & Up.~T. & Low.~T. & Head & Up.~T. & Low.~T.\\
        \midrule
        MOJO-DCT-proj & 76.3 & 102.2 & 99.4 & 86.0 & 105.3 & 83.3\\
        MOJO-proj & {70.3} & {80.7} & {76.7} & {68.3} & {77.7} & {63.2} \\
        \midrule
        MOJO-DCT & 1.32 & 34.0 & 6.97 & {1.32} & {43.1} & {6.97}\\
        MOJO & {1.30} & {32.7} & {6.76} & 1.40 & 44.3 & 7.64\\ \midrule
               Ground truth & \textit{2.17} & \textit{36.4} & \textit{9.78} & \textit{2.52} & \textit{59.1} & \textit{12.5} \\
        \bottomrule
    \end{tabular}
    \vspace{-0.1in}
   \caption{Deformations of body parts (head, Upper Torso, Lower Torso). Scores are in millimeter.
   High values indicate distorted bodies. }
    \label{tab:deformation}
\end{table}

\begin{table}
    \scriptsize
    \centering
    \begin{tabular}{lcccc}
        \toprule
          & \multicolumn{2}{c}{\textbf{ACCAD}~\cite{accad}} & \multicolumn{2}{c}{\textbf{BMLHandball}~\cite{helm2017motion,helm2020integrating}}\\
        \cmidrule(lr){2-3} \cmidrule(lr){4-5}
        Method & foot skate & percep.~score & foot skate & percep.~score\\
        \midrule
        MOJO-DCT & 0.341 & \textbf{4.15}$\pm$1.38 & 0.077 & 4.00$\pm$1.26\\
        MOJO & \textbf{0.278} & 4.07$\pm$\textbf{1.31} & \textbf{0.066} & \textbf{4.17}$\pm$\textbf{1.23}\\
        \midrule
        Ground truth&  \textit{0.067} & \textit{4.82}$\pm$\textit{1.08} & \textit{0.002} & \textit{4.83}$\pm$\textit{1.05}\\
        \bottomrule
    \end{tabular}
    \vspace{-0.1in}
    \caption{Comparison between methods w.r.t. foot skating and the perceptual score, which is given by mean$\pm$std. Best results are in boldface.}
    \label{tab:skating_userstudy}
\end{table}

\begin{table*}[t!]
    \scriptsize
    \centering
    \begin{tabular}{lllllllllllll}
        \toprule
        &\multicolumn{6}{c}{\textbf{ACCAD}~\cite{accad}} & \multicolumn{6}{c}{\textbf{BMLHandball}~\cite{helm2017motion,helm2020integrating}}\\
        \cmidrule(lr){2-7} \cmidrule(lr){8-13}
        Method & \textit{Diversity}$\uparrow$ & \textit{ADE}$\downarrow$ & \textit{FDE}$\downarrow$ & \textit{MMADE}$\downarrow$ & \textit{MMFDE}$\downarrow$ & \textit{BDF}$\downarrow$ & \textit{Diversity}$\uparrow$ & \textit{ADE}$\downarrow$ & \textit{FDE}$\downarrow$ & \textit{MMADE}$\downarrow$ & \textit{MMFDE}$\downarrow$ & \textit{BDF}$\downarrow$ \\
        \midrule
        joints w/o proj. & {21.363} & {1.184} & 2.010 & {1.291} & {2.067} & 0.185 & \textbf{19.091} & {0.930} & {1.132} & {1.000} & {1.156} & 0.205\\
        joints  & 21.106 & 1.192 & 2.022 & 1.299 & 2.076 & 0 & 18.954 & 0.934 & 1.138 & 1.003 & 1.157 & 0 \\
        CMU 41 & 20.676 & {1.214} & {1.919} & 1.306 & 2.080 & {0} & 16.806 & {0.949} & {1.139} & {1.001} & {1.172} & {0}\\
        SSM2 67 & \textbf{24.373} & \textbf{1.124} & \textbf{1.699} & \textbf{1.227} & \textbf{1.838} & 0 & 18.778 & \textbf{0.924} & \textbf{1.099} & \textbf{0.975} & \textbf{1.149} & 0\\
        joints + CMU 41 & 20.988 & 1.187 & 1.841 & 1.308 & 1.967 & 0 & 13.982 & 0.943 & 1.190 & 0.990 & 1.194 & 0\\
        joints + SSM2 67 & 23.504 & 1.166 & 1.892 & 1.276 & 1.953 & 0 & 16.483 & 0.950 & 1.146 & 0.999 & 1.189 & 0\\
        \bottomrule
    \end{tabular}
    \vspace{-0.1in}
    \caption{Comparison between marker-based and joint-based representations. Evaluations are based on the joint locations. {\em BDF} denotes the bone deformation w.r.t. meter. The best results are in boldface. }        
    \vspace{-3mm}
    \label{tab:comparison_jts_markers}
\end{table*}

\myparagraph{Body deformation.}
The results are shown in Tab.~\ref{tab:deformation}.
With the recursive projection scheme,  the body shape is preserved by construction and is close to the ground truth.
Without the projection scheme, the shape of body parts can drift significantly from the true shape, indicated here by high deformation numbers.
MOJO is close to the ground truth but exhibits less deformation suggesting that some nuance is smoothed out by the VAE.


\myparagraph{Foot skating and perceptual score.}
The results are presented in Tab.~\ref{tab:skating_userstudy}.
The model with DCT produces fewer foot skating artifacts, indicating that high-frequency components in the DCT space can better model the foot movements.
In the perceptual study, MOJO performs slightly worse than MOJO-DCT on {\bf ACCAD}, but outperforms it on {\bf BMLhandball}. 
A probable reason is that most actions in {\bf ACCAD} are coarse-grained, whereas most actions in {\bf BMLhandball} are fine-grained. 
The advantage of modelling finer-grained motion of the DCT latent space is more easily perceived in {\bf BMLhandball}.

\subsection{Comparison between Body Representations}
\label{sec:exp:markers_vs_jts}
The body in motion can be represented by locations of joints, markers with different placements, and their combinations. Here we perform a systematic comparison between them.
For the joint-based representation, we use the SMPL~\cite{SMPL:2015} joint locations from {\bf CMU} and {\bf MPI HDM05} to train a CVAE as in MOJO.
A traditional pipeline is to first predict all joints in the future, and then fit the body mesh. Here we also test the performance when applying the recursive body re-projection scheme based on joints.
For fair quantitative evaluation, the metrics are calculated based on the joints of the fitted body meshes.
We re-calculate the diversity, the prediction accuracy metrics, and the eight limb bone deformation (BDF) (according to Eq.~\eqref{eq:deformation} in Appendix) w.r.t.~the joint locations.

\begin{figure}
    \centering
    \includegraphics[width=\linewidth]{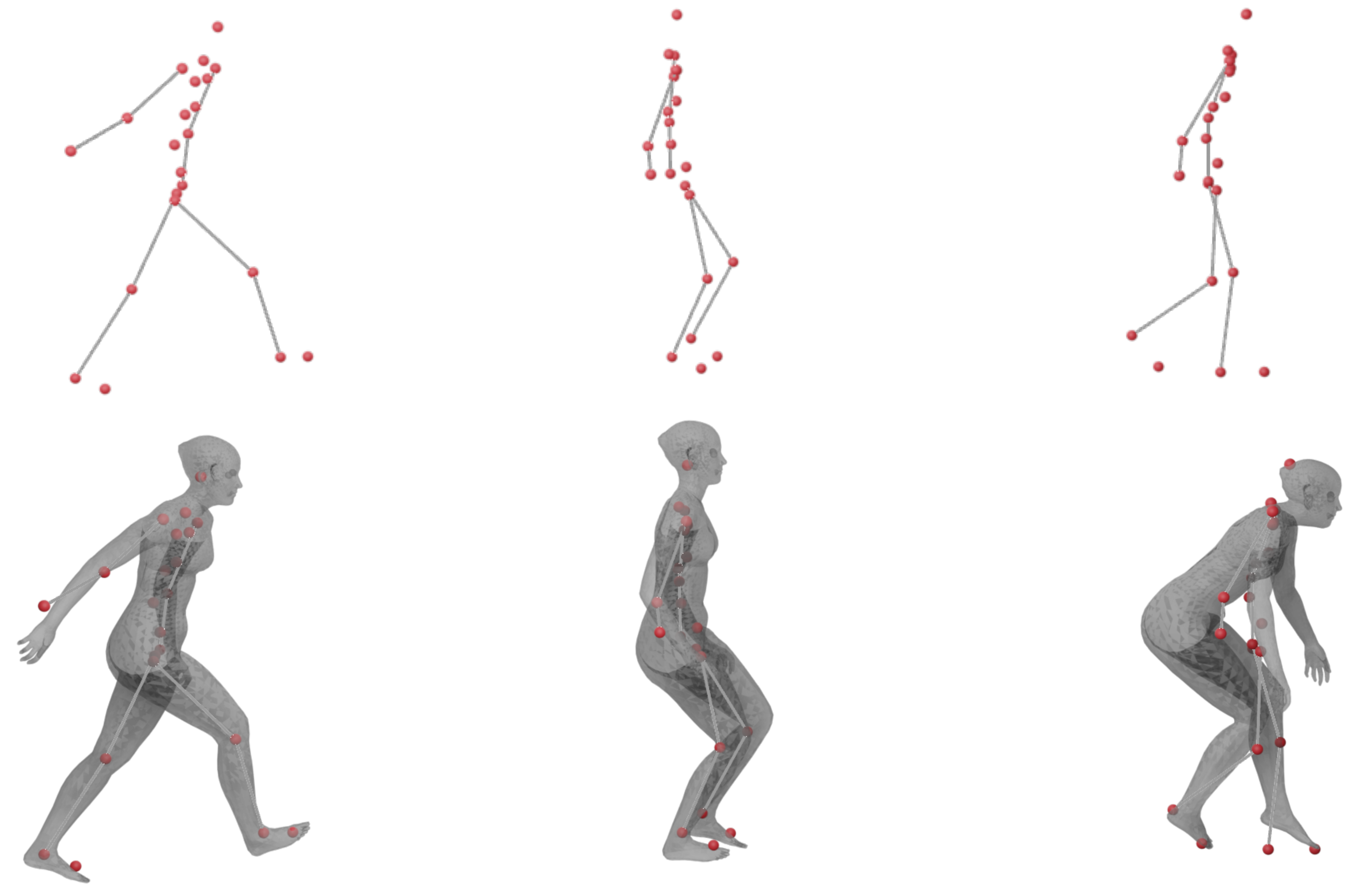}
    \vspace{-0.25in}
    \caption{Fitting a character to predicted joints. The top row and the bottom row show the predicted skeletons and the fitted bodies, respectively. From left to right: The first predicted frame, the middle frame, and the last frame. }
    \vspace{-0.18in}
    \label{fig:fit_jts}
\end{figure}

We randomly choose 60 sequences from each test set and generate 50 future motions based on each sequence. 
Results are presented in Tab.~\ref{tab:comparison_jts_markers}.
The first two rows show that the motion naturalness is improved by the recursive projection, which eliminates the bone deformation. Although the result without projection is slightly better on other measures, the projection scheme completely removes bone deformation and does not produce false poses as in Fig.~\ref{fig:fit_jts}. 
Additionally, using more markers (SSM2 placement with 67 markers) significantly improves performance across the board. 
This shows that the marker distribution is important for motion prediction and that more markers is better.
While MOJO works with markers, joints, or the combination of both, the combination of joints and markers does not produce better performance. 
Note that the joints are never directly observed, but rather are inferred from the markers by commercial mocap systems. Hence, we argue that the joints do not add independent information.

Figure \ref{fig:fit_jts} shows the risk of the traditional joint-based pipeline. 
While the skeletons may look fine to the eye, in the last frame
the character cannot be fit to the joints  due to unrealistic bone lengths.






%% file: sec5-conclusion.tex
\section{Conclusion}
In this paper, we propose MOJO, a new method to predict diverse plausible motions of 3D bodies. 
Instead of using joints to represent the body, MOJO uses a sparse set of markers on the body surface, which better constrain 3D body shape and pose recovery.
In contrast to most existing methods that encode a motion into a single feature vector, we represent motion with latent frequencies, which can describe fine-grained body movements and improve motion prediction consistently.
To produce valid 3D bodies in motion, MOJO uses a recursive projection scheme at test time. By fitting a SMPL-X body to the predicted markers at each frame, the 3D bodies stay valid over time and motion realism is improved. 
Compared to a traditional pipeline based on joints,
 MOJO  thoroughly eliminates implausible body deformations and produces realistic 3D body movements.



Nevertheless, MOJO has some limitations to improve in the future. For example, the recursive projection scheme slows down the inference process.
Also, the motion realism is still not comparable with the ground truth (see Tab.~\ref{tab:skating_userstudy}), indicating room to improve. Moreover, we will explore the performance of MOJO on other marker settings, or even real markers from mocap data.

\vspace{0.1in}
{\small
\noindent\textbf{Acknowledgement:}
We thank Nima Ghorbani for the advice on the body marker setting and the {\bf AMASS} dataset. 
We thank Yinghao Huang, Cornelia K\"{o}hler, Victoria Fern\'{a}ndez Abrevaya, and Qianli Ma for proofreading.
We thank Xinchen Yan and Ye Yuan for discussions on baseline methods.
We thank Shaofei Wang and Siwei Zhang for their help with the user study and the presentation, respectively.

\noindent\textbf{Disclaimer:}
MJB has received research gift funds from Adobe, Intel, Nvidia, Facebook, and Amazon. While MJB is a part-time employee of Amazon, his research was performed solely at, and funded solely by, Max Planck. MJB has financial interests in Amazon Datagen Technologies, and Meshcapade GmbH.
}

%% file: sec6-appendix.tex
\begingroup
\onecolumn 

\appendix
\begin{center}
\Large{\bf We are More than Our Joints: Predicting how 3D Bodies Move \\ **Appendix**}
\end{center}

\setcounter{page}{1}
\setcounter{table}{0}
\setcounter{figure}{0}
\renewcommand{\thetable}{S\arabic{table}}
\renewcommand{\thefigure}{S\arabic{figure}}

MOJO is implemented using SMPL-X but any other parametric 3D body model could be used, e.g.~\cite{STAR:ECCV:2020,xu2020ghum}. To do so, one only needs to implement the recursive fitting of 3D pose to observed markers. This is a straightforward optimization problem. In this paper, we use SMPL-X to demonstrate our MOJO idea, because we can exploit the large-scale AMASS dataset to train/test our networks.
Moreover, SMPL-X is rigged with a skeleton like other body models in computer graphics, so it is completely compatible with standard skeletal techniques.

\section{More Method Details}

\paragraph{Marker placements in our work.}
In our experiments, we use two kinds of marker placements. The first (default) one is the CMU~\cite{mocap_cmu} setting with 41 markers. The second one is the SSM2~\cite{AMASS:ICCV:2019} setting with 67 markers. These marker settings are illustrated in Fig.~\ref{fig:app:markerset}.

\begin{figure}[h!]
    \centering
    \includegraphics[width=0.85\linewidth]{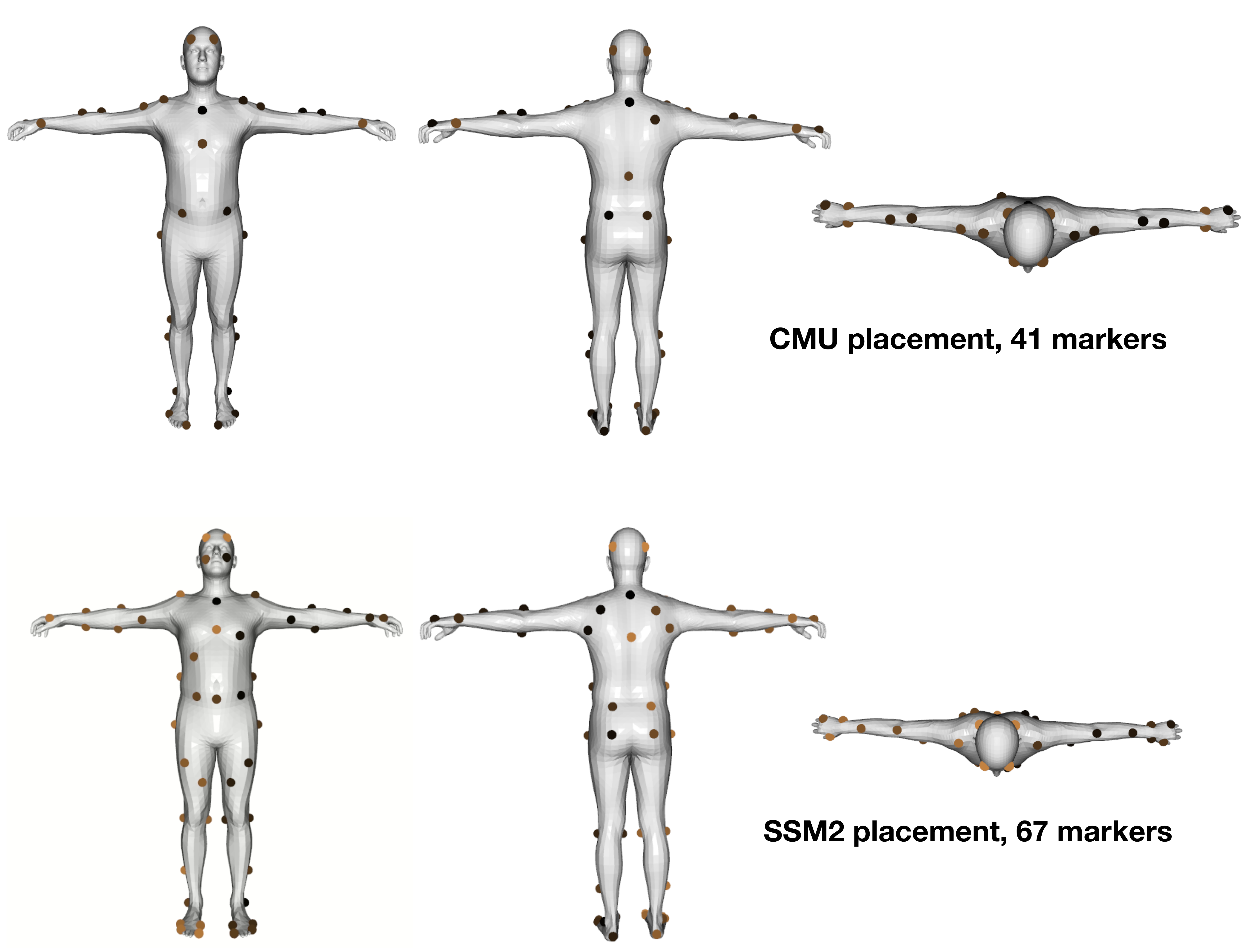}
    \caption{Illustration of our marker settings. The markers are denoted by 3D spheres attached to the SMPL-X body surface. From left to right: the front view, the back view and the top view.}
    \label{fig:app:markerset}
\end{figure}

\paragraph{Network architectures.}
We have demonstrated the CVAE architecture of MOJO in Sec.~\ref{sec:method}, and compare it with several baselines and variants in Sec.~\ref{sec:experiments}. The architectures of the used motion generators are illustrated in Fig.~\ref{fig:app:nets}.
Compared to the CVAE of MOJO, VAE+DCT has no residual connections at the output, and the velocity reconstruction loss is replaced by a loss to minimize $|{\bm x}_M - {\bm y}_0|^2$~\cite{yuan2020dlow}. 
MOJO-DCT-proj encodes the motion ${\bm Y}$ into a single feature vector, rather than a set of frequency components.

\begin{figure}
    \centering
    \includegraphics[width=0.75\linewidth]{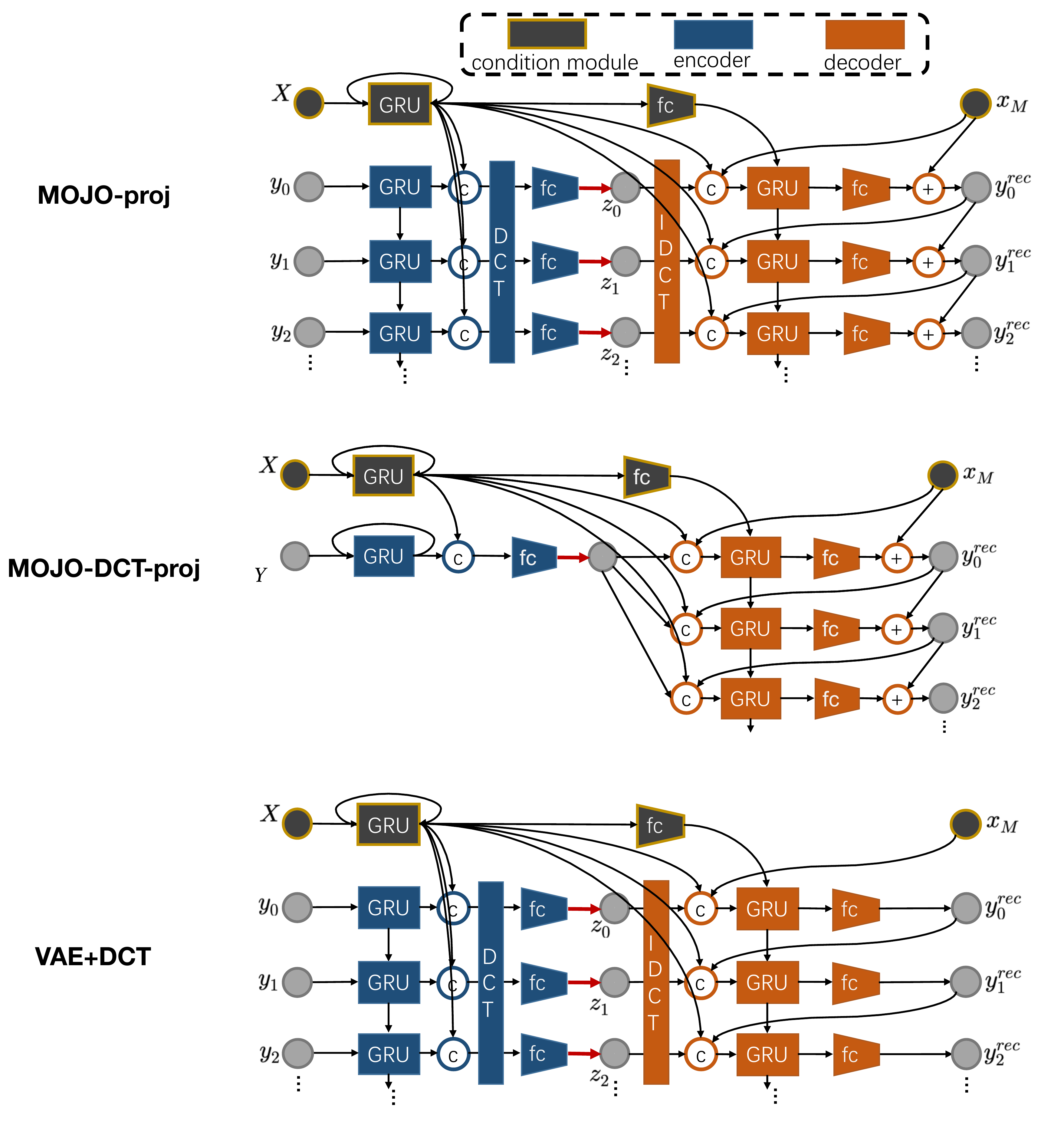}
    \caption{From top to bottom: (1) The CVAE architecture of MOJO. (2) The CVAE architecture of MOJO-DCT, which is used in Tables \ref{tab:marker_compare}~\ref{tab:deformation}~\ref{tab:skating_userstudy}.
    (3) The architecture of VAE+DCT, which is evaluated in Tab.~\ref{tab:skeleton_compare}.
    Note that we only illustrate the motion generators here. The recursive projection scheme can be added during the testing time.}
    \label{fig:app:nets}
\end{figure}

\section{More Experimental Details}

\paragraph{Implementation.}
We use PyTorch v1.6.0~\cite{NEURIPS2019_9015} in our experiments. 
In the training loss Eq.~\eqref{eq:trainloss}, we empirically set $\alpha=3$ in all cases. We find a larger value causes over-smooth and loses motion details, whereas a smaller value can cause jitters. For the models with the latent DCT space, the robust KLD term with the loss weight 1 is employed. For models without the latent DCT space, we use a standard KLD term, and set its weight to 0.1. The weights of the KLD terms are not annealed during training.
In the fitting loss Eq.~\eqref{eq:fitloss}, we empirically set $\{\lambda_1,\lambda_2\}=\{0.0005, 0.01\}$. Smaller values can degrade the pose realism with e.g. twisted torsos, and larger values reduce the pose variations in motion.
Our code is publicly released, which delivers more details.

The model `MOJO-DCT-proj' has a latent dimension of 128. However, `MOJO-proj' suffers from overfitting with the same latent dimension, and hence we set its latent dimension to 16.
We use these dimensions in all experiments.

\paragraph{AMASS sequence canonicalization.}
To unify the sequence length and world coordinates, we canonicalize {\bf AMASS} as follows: \textit{First}, we trim the original sequences into 480-frame (4-second) subsequences, and downsample them from 120fps to 15fps. The condition sequence ${\bm X}$ contains 15 frames (1s) and the future sequence $\bm Y$ contains 45 frames (3s). \textit{Second}, we unify the world coordinates as in \cite{zhang2020perpetual}. For each subsequence, we reset the world coordinate to the SMPL-X~\cite{SMPL-X:2019} body mesh in the first frame: The horizontal X-axis points in the direction from the left hip to the right hip, the Z-axis is the negative direction of gravity, and the Y-axis is pointing forward. The origin is set to to the body's global translation.

\paragraph{More discussions on \textit{MMADE} and \textit{MMFDE}.}
As in~\cite{yuan2020dlow}, we use \textit{MMADE} and \textit{MMFED} to evaluate  prediction accuracy when the input sequence slightly changes. They are regarded as multi-modal version of \textit{ADE} and \textit{FDE}, respectively.
Let's only demonstrate \textit{MMADE} with more details here, since the same principle applies to \textit{MMFDE}.

The \textit{ADE} can be calculated by
\begin{equation}
    e_{\textit{ADE}}(\mathcal{Y}) = \frac{1}{T} \min_{{\bm Y} \in \mathcal{Y}} |{\bm Y} - {\bm Y}_{gt}|^2,
\end{equation}
in which ${\bm Y}$ is a predicted motion, $\mathcal{Y}$ is the set of all predicted motions, and ${\bm Y}_{gt}$ is the ground truth future motion.
In this case, the \textit{MMADE} can be calculate as
\begin{equation}
    e_{\textit{MMADE}}(\mathcal{Y}) = \mathbb{E}_{{\bm Y}^{*} \in \mathcal{Y}_S }   \left[    \frac{1}{T} \min_{{\bm Y} \in \mathcal{Y}} |{\bm Y} - {\bm Y}^{*}|^2 \right]
\end{equation}
with 
\begin{equation}
    \mathcal{Y}_s = \{ {\bm Y}^{*} \in \mathcal{Y}_{gt} \, | \, d({\bm X}^{*}, {\bm X}_{gt}) < \eta\},
\end{equation}
with $\mathcal{Y}_{gt}$ is the set of all ground truth future motion, ${\bm X}$ denotes the corresponding motion in the past, $d(\cdot)$ is a difference measure, and $\eta$ is a pre-defined threshold.
In the work of DLow~\cite{yuan2020dlow}, the difference measure $d(\cdot)$ is based on the L2 difference between the \textit{last frames} of the two motion sequences from the past.

\paragraph{Body deformation metric.}
We measure the markers on the head, the upper torso and the lower torso, respectively.
Specifically, according to {\bf CMU}~\cite{mocap_cmu}, we measure (`LFHD', `RFHD', `RBHD', `LBHD') for the head, (`RSHO', `LSHO', `CLAV', `C7') for the upper torso, and (`RFWT', `LFWT', `LBWT', `RBWT') for the lower torso.
For each rigid body part $P$, the deformation score is the variations of marker pair-wise distances, and is calculated by
\begin{equation}
    s_d(P) = \mathbb{E}_{\bm Y}\left[ \sum_{(i,j)\in P} \sigma_t(|{\bm v}^t_i-{\bm v}_j^t|^2) \right],
    \label{eq:deformation}
\end{equation}
in which ${\bm v}^t_i$ denotes the location of the marker $i$ at time $t$, $\sigma_t$ denotes the standard deviation along the time dimension, and $\mathbb{E}_{\bm Y}[\cdot]$ denotes averaging the scores of different predicted sequences.

\paragraph{User study interface.}
Our user study is performed via AMT. The interface is illustrated in Fig.~\ref{fig:app:userstudy}. We set a six-point Likert scale for evaluation. Each video is evaluated by three subjects.
\begin{figure}
    \centering
    \includegraphics[width=0.75\linewidth]{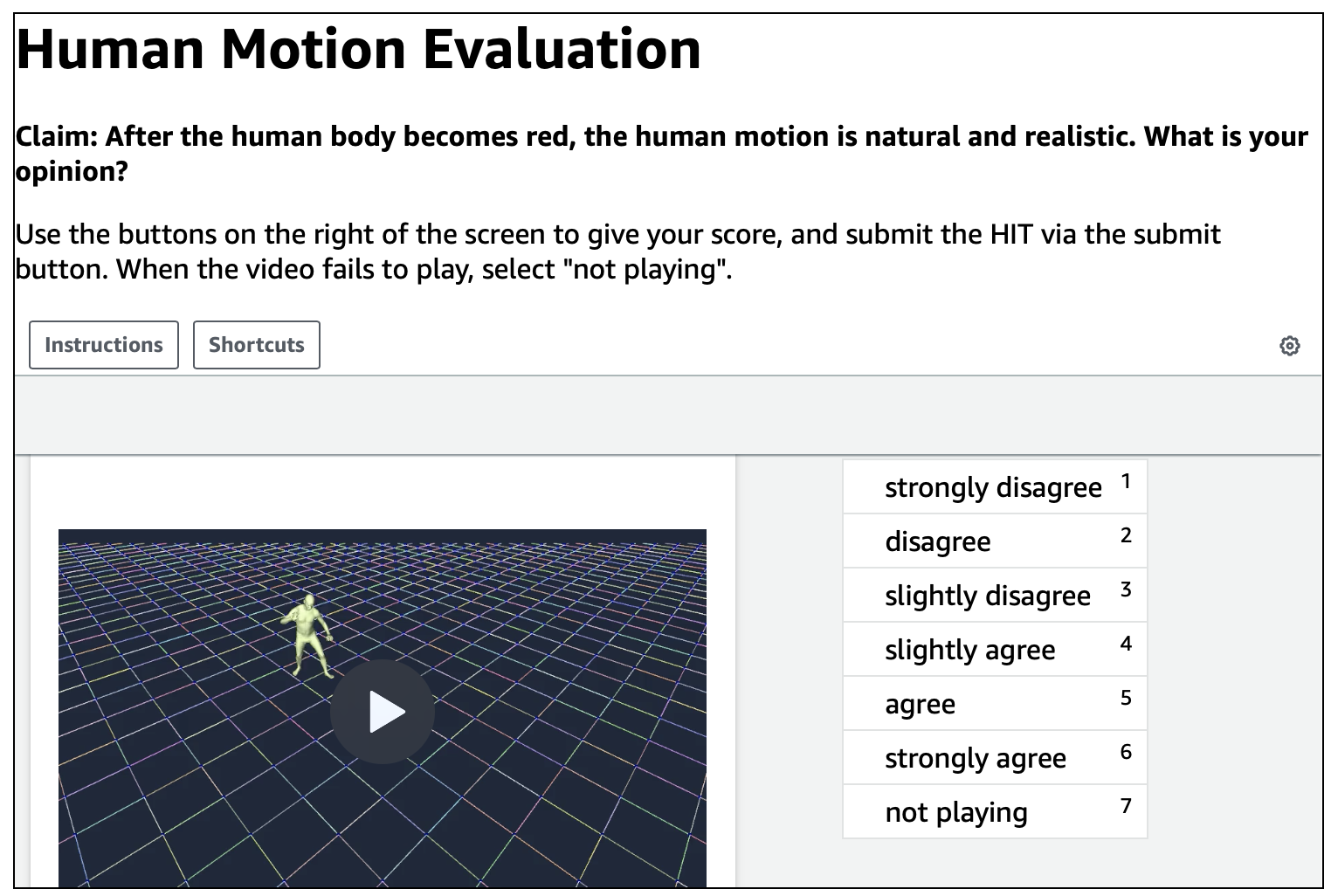}
    \caption{The user interface of our perceptual study on AMT.}
    \label{fig:app:userstudy}
\end{figure}

\paragraph{Performance of the original VAE setting in DLow.}
Noticeably, the DLow CVAE with the original setting cannot directly work on body markers, although it works well with joint locations. Following the evaluation in Tab.~\ref{tab:marker_compare}, the original VAE setting in DLow  gives (diversity=81.10,  ADE=2.79, FDE=4.71, MMADE=2.81, MMFDE=4.71, FSE=0.0031). The diversity is much higher, but the accuracy is considerably worse. Fig.~\ref{fig:app:dlow_vae_fail} shows that its predicted markers are not valid.

\begin{figure}
    \centering
    \includegraphics[width=0.75\linewidth]{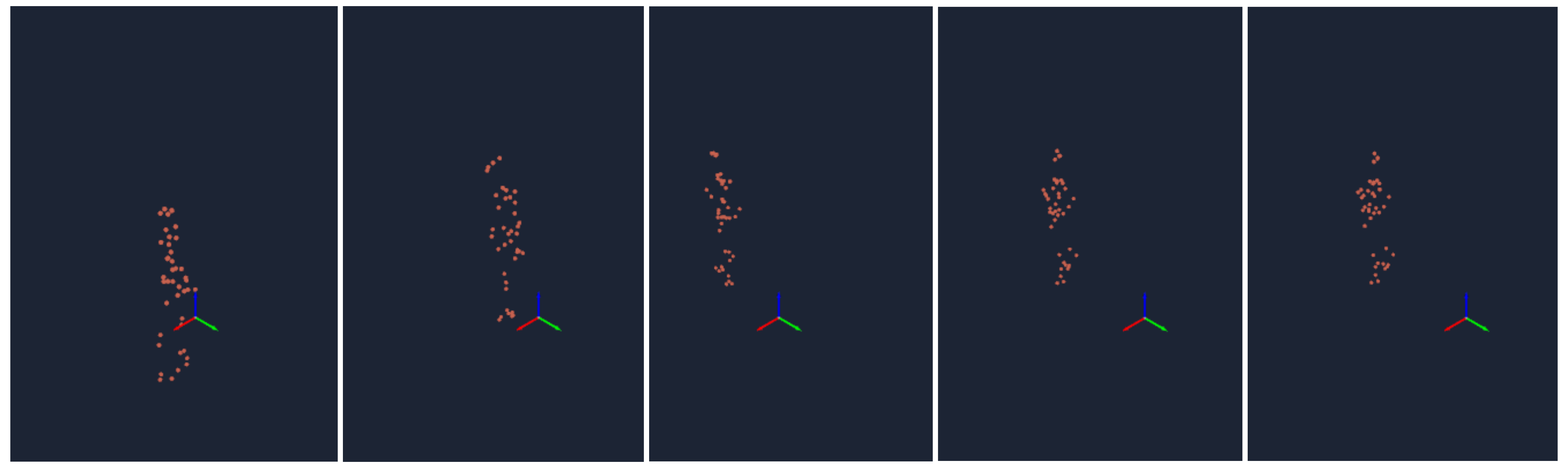}
    \caption{Illustrations of the invalid body markers predicted by the original DLow VAE setting. From left to right are five predicted frames over time.}
    \label{fig:app:dlow_vae_fail}
\end{figure}

\paragraph{Influence of the given frames.}
Based on the CMU markers, we train another two MOJO versions with different input/output sequence lengths. The results are in Tab.~\ref{tab:app:frame_influence}.
As the predicted sequence becomes shorter and the input sequence becomes longer, 
we can see that the accuracy increases but the diversity decreases consistently, indicating that  MOJO  becomes more confident and deterministic.
Such behavior is similar to other state-of-the-art methods like DLow.
Furthermore, to predict even longer motion sequences, one could use a sliding window, and recursively input the predicted sequence to the pre-trained MOJO model to generate new sequences.

\begin{table}[t!]
    \footnotesize
    \centering
    \begin{tabular}{lllllll}
    \toprule
    MOJO frame setting & \textit{Div.}$\uparrow$ & \textit{ADE}$\downarrow$ & \textit{FDE}$\downarrow$ & \textit{MMADE}$\downarrow$ & \textit{MMFDE}$\downarrow$ & \textit{BDF}$\downarrow$ \\
    \midrule
    \multicolumn{7}{c}{\bf ACCAD} \\
    pred.~55 with 5 & \textbf{32.666} & 1.303 & 1.793 & 1.409 & 2.210 & 0\\
    {\it pred.~45 with 15} & 20.676 & {1.214} & {1.919} & 1.306 & 2.080 & {0} \\
    pred.~15 with 45 & 3.728 & \textbf{0.642} & \textbf{1.048} & \textbf{0.701} & \textbf{1.093} & 0\\
    \midrule
    \multicolumn{7}{c}{\bf BMLhandball} \\
    pred.~55 with 5 & \textbf{27.045} & 1.032 & 1.170 & 1.054 & 1.192 & 0 \\
    {\it pred.~45 with 15} & 16.806 & {0.949} & {1.139} & {1.001} & {1.172} & {0} \\
    pred.~15 with 45 & 3.574 & \textbf{0.666} & \textbf{0.993} & \textbf{0.759} & \textbf{1.051} & 0 \\
    \bottomrule
    \end{tabular}
    \caption{Varying the length of the input and output sequence, denoted as `predicting Y frames with X frames'. The setting `{\it pred.~45 with 15}' (predicting 3 sec with 1 sec) was reported in Tab.~\ref{tab:comparison_jts_markers}. Best results are in boldface. }
    \label{tab:app:frame_influence}
\end{table}

\section{More Analysis on the Latent DCT Space}

\paragraph{Analysis on the latent space dimension.}
First, we use {\bf Human3.6M} to analyze the latent DCT space, with joint locations to represent the human body in motion.
Tab. \ref{tab:app:skeleton_latentdim} shows the performance under various settings. Similar to Tab. \ref{tab:skeleton_compare}, as DLow is applied on more frequency bands, the diversity consistently grows, and the motion prediction accuracies are stable. 
Noticeably, VAE+DCT with a 32d latent space outperforms the baseline~\cite{yuan2020dlow} (see Tab. \ref{tab:skeleton_compare}), indicating that our latent DCT space has better representation power.    

\begin{table}
    \centering
    \footnotesize
    \begin{tabular}{llllll}
        \toprule
        Method & \textit{Diversity} & \textit{ADE} & \textit{FDE} & \textit{MMADE} & \textit{MMFDE} \\
        \midrule
        (32d)VAE+DCT & 3.405 & 0.432 & 0.544 & 0.533 & 0.589  \\
        (32d)VAE+DCT+1 & 7.085 & 0.419 & 0.514 & 0.515 & 0.547  \\
        (32d)VAE+DCT+5 & 12.007 & 0.415 & \textbf{0.510} & 0.505 & 0.542 \\
        (32d)VAE+DCT+10 & 13.103 & 0.417 & 0.513 & 0.507 & 0.544 \\
        (32d)VAE+DCT+20 & 14.642 & 0.418 & 0.516 & 0.510 & 0.548  \\
        \midrule
        (64d)VAE+DCT & 3.463 & 0.429 & 0.544 & 0.532 & 0.587  \\
        (64d)VAE+DCT+1 & 7.254 & 0.417 & 0.514 & 0.513 & 0.547  \\
        (64d)VAE+DCT+5 & 12.554 & \textbf{0.413} & \textbf{0.510} & \textbf{0.504} & \textbf{0.540} \\
        (64d)VAE+DCT+10 & 14.233 & 0.414 & 0.514 & 0.506 & 0.546 \\
        (64d)VAE+DCT+20 & \textbf{15.462} & 0.416 & 0.517 & 0.508 & 0.548  \\
        \bottomrule
    \end{tabular}
    \caption{Model performances with different latent dimensions and number of frequency bands with DLow on the {\bf Human3.6M} dataset. Best results are in boldface. This table is directly comparable with Tab.~\ref{tab:skeleton_compare}, which shows the results with the 128d latent space (same with~\cite{yuan2020dlow}).}  
    \label{tab:app:skeleton_latentdim}
\end{table}

Additionally, for the marker-based representation, we evaluate the influence of the latent feature dimension using the {\bf BMLhandball} dataset.
The results are presented in Tab.~\ref{tab:app:marker_ablation}, in which DLow is applied for MOJO-DCT-proj. According to the investigations on the {\bf Human3.6M} dataset, we apply DLow on the lowest 20\% (the lowest 9) frequency bands in MOJO-proj, corresponding to VAE+DCT+20 in Tab.~\ref{tab:app:skeleton_latentdim}.
We can see that the MOJO-DCT-proj performs best with a 128d latent space, yet is worse than most cases of MOJO-proj, which indicates the representation power of the latent DCT space.
In the meanwhile, different versions of MOJO-proj perform comparably with different latent dimensions. As the feature dimension increases, the diversity consistently increases, whereas the prediction accuracies decrease in most cases. 
Therefore, in our experiments, we set the latent dimensions of MOJO-DCT-proj and MOJO-proj to 128 and 16, respectively.

\begin{table}
    \centering
    \footnotesize
    \begin{tabular}{rrllllll}
        \toprule
        Method & \textit{Diversity} & \textit{ADE} & \textit{FDE} & \textit{MMADE} & \textit{MMFDE} & \textit{FSE($10^{-3}$)} \\
        \midrule
        (8d)MOJO-DCT-proj & 0.027 & 2.119 & 3.145 & 2.143 & 3.153 & -2.6 \\
        (16d)MOJO-DCT-proj & 0.060 & 2.105 & 3.134 & 2.125 & 3.133 & -3.5 \\
        (32d)MOJO-DCT-proj & 0.152 & 2.045 & 3.071 & 2.065 & 3.068 & -3.9 \\
        (64d)MOJO-DCT-proj & 17.405 & 1.767 & 2.213 & 1.790 & 2.219 & 0.2 \\
        (128d)MOJO-DCT-proj & \textbf{21.504} & \textbf{1.608} & \textbf{1.914} & \textbf{1.628} & \textbf{1.919} & \textbf{0.0} \\
        \midrule
        (8d)MOJO-proj & 20.236 & \textbf{1.525} & 1.893 & 1.552 & 1.893 & -0.8 \\
        (16d)MOJO-proj & 23.660 & 1.528 & 1.848 & \textbf{1.550} & 1.847 & 0.4 \\
        (32d)MOJO-proj & 24.448 & 1.554 & 1.850 & 1.573 & 1.846 & 1.0 \\
        (64d)MOJO-proj & 24.129 & 1.557 & \textbf{1.820} & 1.576 & \textbf{1.819} & 1.0 \\
        (128d)MOJO-proj & \textbf{25.265} & 1.620 & 1.852 & 1.636 & 1.851 & \textbf{2.5} \\
        \bottomrule
    \end{tabular}
    \caption{Performances with various latent space dimensions on {\bf BMLhandball}. Best results of each model are in boldface. }
    \label{tab:app:marker_ablation}
\end{table}

\paragraph{Visualization of the latent DCT space.}
Since the latent space is in the frequency domain, we visualize the average frequency spectra of the inference posterior in Fig. \ref{fig:app:skeleton_spectra}, based on the VAE+DCT(128d) model and the {\bf Human3.6M} dataset. 
We find that the bias of fc layers between the DCT and the inverse DCT can lead to stair-like structures. For both cases with and without the fc layer bias, we can observe that most information is concentrated at low-frequency bands. This fact can explain the performance saturation when employing DLow on more frequency bands, and also fits the energy compaction property of DCT. 
Moreover, we show the performance without the fc bias in Tab. \ref{tab:app:skeleton_fcbias}. Compared to the results in Tab. \ref{tab:skeleton_compare}, we find that the influence of these bias values is trivial. 
Therefore, in our experiments we preserve the bias values in these fc layers trainable.

\begin{table}
    \centering
    \footnotesize
    \begin{tabular}{llllll}
        \toprule
        Method & \textit{Diversity} & \textit{ADE} & \textit{FDE} & \textit{MMADE} & \textit{MMFDE} \\
        \midrule
        VAE+DCT-fcbias & 3.442 & 0.431 & 0.547 & 0.525 & 0.584  \\
        VAE+DCT+1-fcbias & 7.072 & 0.417 & 0.514 & 0.506 & 0.541  \\
        VAE+DCT+5-fcbias & 13.051 & \textbf{0.413} & \textbf{0.512} & \textbf{0.498} & \textbf{0.537} \\
        VAE+DCT+10-fcbias & 14.723 & 0.415 & 0.515 & 0.500 & 0.540 \\
        VAE+DCT+20-fcbias & \textbf{16.008} & 0.415 & 0.517 & 0.501 & 0.542  \\
        \bottomrule
    \end{tabular}
    \caption{The model performances with zero values of the fc layer bias. `-fcbias' denotes no bias. The best results are in boldface, and can be directly compared with the results in Tab. \ref{tab:skeleton_compare}.}
    \label{tab:app:skeleton_fcbias}
\end{table}

\begin{figure}
    \centering
    \includegraphics[width=0.8\linewidth]{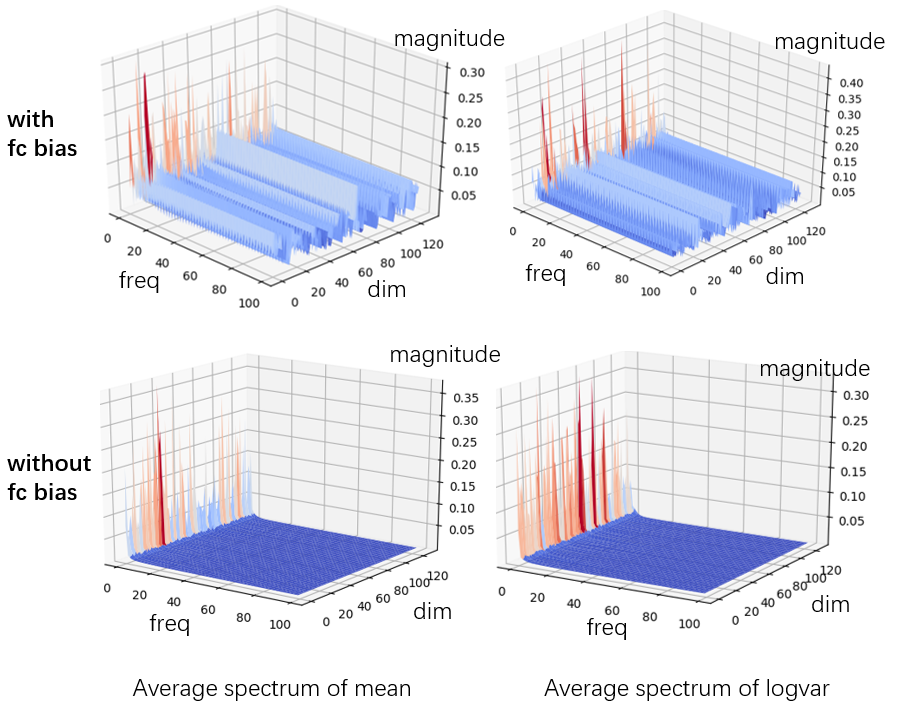}
    \caption{With the VAE+DCT(128d) model, we randomly select 5000 samples from the {\bf Human3.6M} training set, and obtain their mean and the logarithmic variance values from the VAE encoder. To show the frequency spectra, we average all absolute values of mean or logarithmic variance. Note that when both mean and logarithmic variance are zero, the posterior is equal to the standard normal distribution, which only produces white noise without information.}
    \label{fig:app:skeleton_spectra}
\end{figure}

%% file: main.bbl
\begin{thebibliography}{10}\itemsep=-1pt

\bibitem{mocap_cmu}
{CMU} graphics lab. {CMU} graphics lab motion capture database.
\newblock \url{http://mocap.cs.cmu.edu/}, 2000.

\bibitem{accad}
{ACCAD MoCap System and Data}.
\newblock \url{https://accad.osu.edu/research/motion-lab/systemdata}, 2018.

\bibitem{ahmed1974discrete}
Nasir Ahmed, T. Natarajan, and Kamisetty~R. Rao.
\newblock Discrete cosine transform.
\newblock {\em IEEE transactions on Computers}, 100(1):90--93, 1974.

\bibitem{akhter2012bilinear}
Ijaz Akhter, Tomas Simon, Sohaib Khan, Iain Matthews, and Yaser Sheikh.
\newblock Bilinear spatiotemporal basis models.
\newblock {\em ACM Transactions on Graphics}, 31(2):1--12, 2012.

\bibitem{aksan2020attention}
Emre Aksan, Peng Cao, Manuel Kaufmann, and Otmar Hilliges.
\newblock Attention, please: A spatio-temporal transformer for {3D} human
  motion prediction.
\newblock {\em arXiv preprint arXiv:2004.08692}, 2020.

\bibitem{aksan2019structured}
Emre Aksan, Manuel Kaufmann, and Otmar Hilliges.
\newblock Structured prediction helps {3D} human motion modelling.
\newblock In {\em IEEE Conference on Computer Vision and Pattern Recognition},
  pages 7144--7153, 2019.

\bibitem{barsoum2018hp}
Emad Barsoum, John Kender, and Zicheng Liu.
\newblock {HP-GAN}: Probabilistic {3D} human motion prediction via {GAN}.
\newblock In {\em IEEE Conf. Comput. Vis. Pattern Recog. Worksh.}, pages
  1418--1427, 2018.

\bibitem{bhattacharyya2018accurate}
Apratim Bhattacharyya, Bernt Schiele, and Mario Fritz.
\newblock Accurate and diverse sampling of sequences based on a “best of
  many” sample objective.
\newblock In {\em IEEE Conference on Computer Vision and Pattern Recognition},
  pages 8485--8493, 2018.

\bibitem{Bogo:ECCV:2016}
Federica Bogo, Angjoo Kanazawa, Christoph Lassner, Peter Gehler, Javier Romero,
  and Michael~J. Black.
\newblock Keep it {SMPL}: Automatic estimation of {3D} human pose and shape
  from a single image.
\newblock In {\em European Conference on Computer Vision}, Oct. 2016.

\bibitem{cai2020learning}
Yujun Cai, Lin Huang, Yiwei Wang, Tat-Jen Cham, Jianfei Cai, Junsong Yuan, Jun
  Liu, Xu Yang, Yiheng Zhu, Xiaohui Shen, et~al.
\newblock Learning progressive joint propagation for human motion prediction.
\newblock In {\em European Conference on Computer Vision}, 2020.

\bibitem{charbonnier1994two}
Pierre Charbonnier, Laure Blanc-Feraud, Gilles Aubert, and Michel Barlaud.
\newblock Two deterministic half-quadratic regularization algorithms for
  computed imaging.
\newblock In {\em Proceedings of the International Conference on Image
  Processing}, pages 168--172, 1994.

\bibitem{cho-etal-2014-learning}
Kyunghyun Cho, Bart van Merri{\"e}nboer, Caglar Gulcehre, Dzmitry Bahdanau,
  Fethi Bougares, Holger Schwenk, and Yoshua Bengio.
\newblock Learning phrase representations using {RNN} encoder{--}decoder for
  statistical machine translation.
\newblock In {\em Proceedings of the Conference on Empirical Methods in Natural
  Language Processing}, pages 1724--1734, Doha, Qatar, Oct. 2014.

\bibitem{cui2020learning}
Qiongjie Cui, Huaijiang Sun, and Fei Yang.
\newblock Learning dynamic relationships for {3D} human motion prediction.
\newblock In {\em IEEE Conference on Computer Vision and Pattern Recognition},
  pages 6519--6527, 2020.

\bibitem{dilokthanakul2016deep}
Nat Dilokthanakul, Pedro~AM Mediano, Marta Garnelo, Matthew~CH Lee, Hugh
  Salimbeni, Kai Arulkumaran, and Murray Shanahan.
\newblock Deep unsupervised clustering with gaussian mixture variational
  autoencoders.
\newblock {\em arXiv preprint arXiv:1611.02648}, 2016.

\bibitem{ghosh2017learning}
Partha Ghosh, Jie Song, Emre Aksan, and Otmar Hilliges.
\newblock Learning human motion models for long-term predictions.
\newblock In {\em International Conference on 3D Vision}, pages 458--466. IEEE,
  2017.

\bibitem{gopalakrishnan2019neural}
Anand Gopalakrishnan, Ankur Mali, Dan Kifer, Lee Giles, and Alexander~G
  Ororbia.
\newblock A neural temporal model for human motion prediction.
\newblock In {\em IEEE Conference on Computer Vision and Pattern Recognition},
  pages 12116--12125, 2019.

\bibitem{gui2018adversarial}
Liang-Yan Gui, Yu-Xiong Wang, Xiaodan Liang, and Jos{\'e}~MF Moura.
\newblock Adversarial geometry-aware human motion prediction.
\newblock In {\em European Conference on Computer Vision}, pages 786--803,
  2018.

\bibitem{gui2018few}
Liang-Yan Gui, Yu-Xiong Wang, Deva Ramanan, and Jos{\'e}~MF Moura.
\newblock Few-shot human motion prediction via meta-learning.
\newblock In {\em European Conference on Computer Vision}, pages 432--450,
  2018.

\bibitem{gurumurthy2017deligan}
Swaminathan Gurumurthy, Ravi Kiran~Sarvadevabhatla, and R Venkatesh~Babu.
\newblock {DeLiGAN}: Generative adversarial networks for diverse and limited
  data.
\newblock In {\em IEEE Conference on Computer Vision and Pattern Recognition},
  pages 166--174, 2017.

\bibitem{helm2020integrating}
Fabian Helm, Rouwen Ca{\~n}al-Bruland, David~L Mann, Nikolaus~F Troje, and
  J{\"o}rn Munzert.
\newblock Integrating situational probability and kinematic information when
  anticipating disguised movements.
\newblock {\em Psychology of Sport and Exercise}, 46:101607, 2020.

\bibitem{helm2017motion}
Fabian Helm, Nikolaus~F Troje, and J{\"o}rn Munzert.
\newblock Motion database of disguised and non-disguised team handball penalty
  throws by novice and expert performers.
\newblock {\em Data in brief}, 15:981--986, 2017.

\bibitem{hernandez2019human}
Alejandro Hernandez, Jurgen Gall, and Francesc Moreno-Noguer.
\newblock Human motion prediction via spatio-temporal inpainting.
\newblock In {\em IEEE Conference on Computer Vision and Pattern Recognition},
  pages 7134--7143, 2019.

\bibitem{holden2017fast}
Daniel Holden, Ikhsanul Habibie, Ikuo Kusajima, and Taku Komura.
\newblock Fast neural style transfer for motion data.
\newblock {\em IEEE computer graphics and applications}, 37(4):42--49, 2017.

\bibitem{huang2017towards}
Yinghao Huang, Federica Bogo, Christoph Lassner, Angjoo Kanazawa, Peter~V
  Gehler, Javier Romero, Ijaz Akhter, and Michael~J Black.
\newblock Towards accurate marker-less human shape and pose estimation over
  time.
\newblock In {\em International Conference on 3D Vision}, pages 421--430. IEEE,
  2017.

\bibitem{h36m_pami}
Catalin Ionescu, Dragos Papava, Vlad Olaru, and Cristian Sminchisescu.
\newblock {Human3.6M}: Large scale datasets and predictive methods for {3D}
  human sensing in natural environments.
\newblock {\em IEEE Transactions on Pattern Analysis and Machine intelligence},
  36(7):1325--1339, 2014.

\bibitem{kingma2013auto}
Diederik~P Kingma and Max Welling.
\newblock Auto-encoding variational {Bayes}.
\newblock In {\em International Conference on Learning Representions}, 2014.

\bibitem{VIBE:CVPR:2020}
Muhammed Kocabas, Nikos Athanasiou, and Michael~J. Black.
\newblock {VIBE}: Video inference for human body pose and shape estimation.
\newblock In {\em IEEE Conference on Computer Vision and Pattern Recognition},
  pages 5252--5262, 2020.

\bibitem{li2018convolutional}
Chen Li, Zhen Zhang, Wee Sun~Lee, and Gim Hee~Lee.
\newblock Convolutional sequence to sequence model for human dynamics.
\newblock In {\em IEEE Conference on Computer Vision and Pattern Recognition},
  pages 5226--5234, 2018.

\bibitem{li2020dynamic}
Maosen Li, Siheng Chen, Yangheng Zhao, Ya Zhang, Yanfeng Wang, and Qi Tian.
\newblock Dynamic multiscale graph neural networks for {3D} skeleton based
  human motion prediction.
\newblock In {\em IEEE Conference on Computer Vision and Pattern Recognition},
  pages 214--223, 2020.

\bibitem{ling2020character}
Hung~Yu Ling, Fabio Zinno, George Cheng, and Michiel Van De~Panne.
\newblock Character controllers using motion {VAEs}.
\newblock {\em ACM Transactions on Graphics}, 39(4):40--1, 2020.

\bibitem{liu2017human}
Hongyi Liu and Lihui Wang.
\newblock Human motion prediction for human-robot collaboration.
\newblock {\em Journal of Manufacturing Systems}, 44:287--294, 2017.

\bibitem{loper2014mosh}
Matthew Loper, Naureen Mahmood, and Michael~J Black.
\newblock Mosh: Motion and shape capture from sparse markers.
\newblock {\em ACM Transactions on Graphics}, 33(6):1--13, 2014.

\bibitem{SMPL:2015}
Matthew Loper, Naureen Mahmood, Javier Romero, Gerard Pons-Moll, and Michael~J.
  Black.
\newblock {SMPL}: A skinned multi-person linear model.
\newblock {\em ACM Transactions on Graphics}, 34(6):248:1--248:16, Oct. 2015.

\bibitem{AMASS:ICCV:2019}
Naureen Mahmood, Nima Ghorbani, Nikolaus~F. Troje, Gerard Pons-Moll, and
  Michael~J. Black.
\newblock {AMASS}: Archive of motion capture as surface shapes.
\newblock In {\em International Conference on Computer Vision}, pages
  5442--5451, Oct. 2019.

\bibitem{mao2019learning}
Wei Mao, Miaomiao Liu, Mathieu Salzmann, and Hongdong Li.
\newblock Learning trajectory dependencies for human motion prediction.
\newblock In {\em International Conference on Computer Vision}, pages
  9489--9497, 2019.

\bibitem{martinez2017human}
Julieta Martinez, Michael~J Black, and Javier Romero.
\newblock On human motion prediction using recurrent neural networks.
\newblock In {\em IEEE Conference on Computer Vision and Pattern Recognition},
  pages 2891--2900, 2017.

\bibitem{muller2007documentation}
Meinard M{\"u}ller, Tido R{\"o}der, Michael Clausen, Bernhard Eberhardt,
  Bj{\"o}rn Kr{\"u}ger, and Andreas Weber.
\newblock Documentation mocap database {HDM05}.
\newblock 2007.

\bibitem{ormoneit2000learning}
Dirk Ormoneit, Hedvig Sidenbladh, Michael Black, and Trevor Hastie.
\newblock Learning and tracking cyclic human motion.
\newblock {\em Advances in Neural Information Processing Systems}, 13:894--900,
  2000.

\bibitem{STAR:ECCV:2020}
Ahmed A.~A. Osman, Timo Bolkart, and Michael~J. Black.
\newblock {STAR}: Sparse trained articulated human body regressor.
\newblock In {\em European Conference on Computer Vision}, volume LNCS 12355,
  pages 598--613, Aug. 2020.

\bibitem{NEURIPS2019_9015}
Adam Paszke, Sam Gross, Francisco Massa, Adam Lerer, James Bradbury, Gregory
  Chanan, Trevor Killeen, Zeming Lin, Natalia Gimelshein, Luca Antiga, Alban
  Desmaison, Andreas Kopf, Edward Yang, Zachary DeVito, Martin Raison, Alykhan
  Tejani, Sasank Chilamkurthy, Benoit Steiner, Lu Fang, Junjie Bai, and Soumith
  Chintala.
\newblock Pytorch: An imperative style, high-performance deep learning library.
\newblock In H. Wallach, H. Larochelle, A. Beygelzimer, F. d\textquotesingle
  Alch\'{e}-Buc, E. Fox, and R. Garnett, editors, {\em Advances in Neural
  Information Processing Systems 32}, pages 8024--8035. Curran Associates,
  Inc., 2019.

\bibitem{SMPL-X:2019}
Georgios Pavlakos, Vasileios Choutas, Nima Ghorbani, Timo Bolkart, Ahmed A.~A.
  Osman, Dimitrios Tzionas, and Michael~J. Black.
\newblock Expressive body capture: {3D} hands, face, and body from a single
  image.
\newblock In {\em IEEE Conference on Computer Vision and Pattern Recognition},
  pages 10975--10985, June 2019.

\bibitem{pavllo2019modeling}
Dario Pavllo, Christoph Feichtenhofer, Michael Auli, and David Grangier.
\newblock Modeling human motion with quaternion-based neural networks.
\newblock {\em International Journal of Computer Vision}, pages 1--18, 2019.

\bibitem{pavllo2018quaternet}
Dario Pavllo, David Grangier, and Michael Auli.
\newblock Quaternet: A quaternion-based recurrent model for human motion.
\newblock In {\em British Machine Vision Conference}, 2018.

\bibitem{889031}
K. {Pullen} and C. {Bregler}.
\newblock Animating by multi-level sampling.
\newblock In {\em Proceedings Computer Animation 2000}, pages 36--42, 2000.

\bibitem{rao2014discrete}
K~Ramamohan Rao and Ping Yip.
\newblock {\em Discrete cosine transform: algorithms, advantages,
  applications}.
\newblock Academic press, 2014.

\bibitem{MANO:SIGGRAPHASIA:2017}
Javier Romero, Dimitrios Tzionas, and Michael~J. Black.
\newblock {Embodied Hands}: Modeling and capturing hands and bodies together.
\newblock {\em ACM Transactions on Graphics, (Proc. SIGGRAPH Asia)}, 36(6),
  Nov. 2017.

\bibitem{rudenko2020human}
Andrey Rudenko, Luigi Palmieri, Michael Herman, Kris~M Kitani, Dariu~M Gavrila,
  and Kai~O Arras.
\newblock Human motion trajectory prediction: A survey.
\newblock {\em The International Journal of Robotics Research}, 39(8):895--935,
  2020.

\bibitem{sigal2010humaneva}
Leonid Sigal, Alexandru~O Balan, and Michael~J Black.
\newblock {HumanEva}: Synchronized video and motion capture dataset and
  baseline algorithm for evaluation of articulated human motion.
\newblock {\em International Journal of Computer Vision}, 87(1-2):4, 2010.

\bibitem{tang2018long}
Yongyi Tang, Lin Ma, Wei Liu, and Wei-Shi Zheng.
\newblock Long-term human motion prediction by modeling motion context and
  enhancing motion dynamic.
\newblock In {\em International Joint Conference on Artificial Intelligence},
  page 935–941, 2018.

\bibitem{walker2017pose}
Jacob Walker, Kenneth Marino, Abhinav Gupta, and Martial Hebert.
\newblock The pose knows: Video forecasting by generating pose futures.
\newblock In {\em International Conference on Computer Vision}, pages
  3332--3341, 2017.

\bibitem{wei2020his}
Mao Wei, Liu Miaomiao, and Salzemann Mathieu.
\newblock History repeats itself: Human motion prediction via motion attention.
\newblock In {\em European Conference on Computer Vision}, 2020.

\bibitem{Weng2020_SPF2_eccvw}
Xinshuo Weng, Jianren Wang, Sergey Levine, Kris Kitani, and Nick Rhinehart.
\newblock {4D Forecasting: Sequantial Forecasting of 100,000 Points}.
\newblock {\em Euro. Conf. Comput. Vis. Worksh.}, 2020.

\bibitem{Weng2020_SPF2}
Xinshuo Weng, Jianren Wang, Sergey Levine, Kris Kitani, and Nick Rhinehart.
\newblock {Inverting the Pose Forecasting Pipeline with SPF2: Sequential
  Pointcloud Forecasting for Sequential Pose Forecasting}.
\newblock {\em CoRL}, 2020.

\bibitem{wu2020futurepong}
Erwin Wu and Hideki Koike.
\newblock Futurepong: Real-time table tennis trajectory forecasting using pose
  prediction network.
\newblock In {\em Extended Abstracts of the 2020 CHI Conference on Human
  Factors in Computing Systems}, pages 1--8, 2020.

\bibitem{xu2020ghum}
Hongyi Xu, Eduard~Gabriel Bazavan, Andrei Zanfir, William~T Freeman, Rahul
  Sukthankar, and Cristian Sminchisescu.
\newblock {GHUM} \& {GHUML}: Generative {3D} human shape and articulated pose
  models.
\newblock In {\em IEEE Conference on Computer Vision and Pattern Recognition},
  pages 6184--6193, 2020.

\bibitem{yan2018mt}
Xinchen Yan, Akash Rastogi, Ruben Villegas, Kalyan Sunkavalli, Eli Shechtman,
  Sunil Hadap, Ersin Yumer, and Honglak Lee.
\newblock {MT-VAE}: Learning motion transformations to generate multimodal
  human dynamics.
\newblock In {\em European Conference on Computer Vision}, pages 265--281,
  2018.

\bibitem{yuan2020dlow}
Ye Yuan and Kris Kitani.
\newblock {DLow}: Diversifying latent flows for diverse human motion
  prediction.
\newblock {\em European Conference on Computer Vision}, 2020.

\bibitem{yuan2019diverse}
Ye Yuan and Kris~M. Kitani.
\newblock Diverse trajectory forecasting with determinantal point processes.
\newblock In {\em International Conference on Learning Representions}, 2020.

\bibitem{yumer2016spectral}
M~Ersin Yumer and Niloy~J Mitra.
\newblock Spectral style transfer for human motion between independent actions.
\newblock {\em ACM Transactions on Graphics}, 35(4):1--8, 2016.

\bibitem{zhang2019predicting}
Jason~Y Zhang, Panna Felsen, Angjoo Kanazawa, and Jitendra Malik.
\newblock Predicting {3D} human dynamics from video.
\newblock In {\em International Conference on Computer Vision}, pages
  7114--7123, 2019.

\bibitem{zhang2020perpetual}
Yan Zhang, Michael~J Black, and Siyu Tang.
\newblock Perpetual motion: Generating unbounded human motion.
\newblock {\em arXiv preprint arXiv:2007.13886}, 2020.

\bibitem{zhou2019continuity}
Yi Zhou, Connelly Barnes, Jingwan Lu, Jimei Yang, and Hao Li.
\newblock On the continuity of rotation representations in neural networks.
\newblock In {\em IEEE Conference on Computer Vision and Pattern Recognition},
  pages 5745--5753, 2019.

\bibitem{zhou2018autoconditioned}
Yi Zhou, Zimo Li, Shuangjiu Xiao, Chong He, Zeng Huang, and Hao Li.
\newblock Auto-conditioned recurrent networks for extended complex human motion
  synthesis.
\newblock In {\em International Conference on Learning Representions}, 2018.

\end{thebibliography}
